\documentclass[letterpaper]{article} 
\usepackage{amsmath,amsthm,amsfonts} 
\usepackage{aaai2026}  
\usepackage{times}  
\usepackage{helvet}  
\usepackage{courier}  
\usepackage[hyphens]{url}  
\usepackage{graphicx} 
\usepackage{natbib}  
\usepackage{caption} 
\usepackage{algorithm}
\usepackage{algorithmic}
\usepackage{multirow}
\let\checkmark\relax 
\usepackage{dingbat}
\usepackage{color}
\usepackage[table]{colortbl}
\usepackage{arydshln}
\usepackage{graphicx}
\usepackage{subcaption}

\usepackage{fancyhdr}
\usepackage{newfloat}
\usepackage{listings}
\DeclareCaptionStyle{ruled}{labelfont=normalfont,labelsep=colon,strut=off} 
\floatstyle{ruled}
\newfloat{listing}{tb}{lst}{}
\floatname{listing}{Listing}
\title{MoSSDA: A Semi-Supervised Domain Adaptation Framework for Multivariate Time-Series Classification using Momentum Encoder}
\author {
    Seonyoung Kim\textsuperscript{\rm 1},
    Dongil Kim\textsuperscript{\rm 2}\thanks{Corresponding author.}   
}
\affiliations {
    \textsuperscript{\rm 1}Department of Artificial Intelligence and Software, Ewha Womans University\\
    \textsuperscript{\rm 2}Department of Data Science, Ewha Womans University\\
    1020seonyoung@ewha.ac.kr, d.kim@ewha.ac.kr
}

\usepackage{bibentry}

\begin{document}

\maketitle

\begin{abstract}
Deep learning has emerged as the most promising approach in various fields; however, when the distributions of training and test data are different (domain shift), the performance of deep learning models can degrade. 
Semi-supervised domain adaptation (SSDA) is a major approach for addressing this issue, assuming that a fully labeled training set (source domain) is available, but the test set (target domain) provides labels only for a small subset.
In this study, we propose a novel two-step momentum encoder-utilized SSDA framework, MoSSDA, for multivariate time-series classification.
Time series data are highly sensitive to noise, and sequential dependencies cause domain shifts resulting in critical performance degradation. To obtain a robust, domain-invariant and class-discriminative representation, MoSSDA employs a domain-invariant encoder to learn features from both source and target domains. Subsequently, the learned features are fed to a mixup-enhanced positive contrastive module consisting of an online momentum encoder. The final classifier is trained with learned features that exhibit consistency and discriminability with limited labeled target domain data, without data augmentation.
We applied a two-stage process by separating the gradient flow between the encoders and the classifier to obtain rich and complex representations.
Through extensive experiments on six diverse datasets, MoSSDA achieved state-of-the-art performance for three different backbones and various unlabeled ratios in the target domain data. The Ablation study confirms that each module, including two-stage learning, is effective in improving the performance. Our code is available at https://github.com/seonyoungKimm/MoSSDA
\end{abstract}


\section{Introduction} \label{sec:introduction}
The advent of deep learning has led numerous models demonstrating remarkable performance across various domains. Specifically, time-series classification has become a significant and challenging problem in various applications, including medicine, manufacturing, and human activity recognition \cite{eldele2021time2,chang2020systematic1,li2021modeling3,ragab2023adatime,deng2021multiUDAecg}.
Time-series data require a different approach compared to other data types because their continuous nature includes temporal dependencies, trends, and recurring patterns. In the case of multivariate time series, the data becomes even more complex because of the intermingling of channel dynamics and channel-dependency information. These inherent characteristics make multivariate time-series classification particularly challenging.

Real-world time-series data are prone to variations owing to factors, such as collection environment, sensor type, and recording conditions.
Therefore, time-series data often exhibit significant shifts in their distribution. This phenomenon, termed "domain shift," violates the fundamental independent and identically distributed (i.i.d.) assumption underlying numerous machine learning models \cite{ott2022domainLabelShift}.
However, deep-learning-based time-series models tend to degrade their performance when the test data  distribution (target domain) differ from that of the training data (source domain).

The domain adaptation approach aims to address this challenge and has gained interest from deep learning and time-series researchers.
Two main research directions exist in domain adaptation: unsupervised and semi-supervised domain adaptation (SSDA). Unsupervised domain adaptation (UDA) methods presume the complete absence of labels in the target domain. However, many practical scenarios allow for the acquisition of a limited valuable set of labeled target domain data. In such settings, which are addressed by SSDA, a pragmatic approach is to leverage a small set of labeled target instances alongside a larger corpus of unlabeled data. This strategy was demonstrated to be effective in resolving the distribution discrepancy between the source and target domains \cite{saito2019minimaxSSDA,kim2022DARK}.

Data augmentation has emerged as a prominent strategy for utilizing limited target domain information, largely motivated by its profound success in computer vision \cite{ilbert2024augTS2,iglesias2023augTS1}.
However, because temporal order and sequential dependencies are crucial in time-series data, the common transformation-based augmentation used in spatial data, such as rotation or random cropping, may disrupt critical temporal characteristics \cite{chang2024timedrl} and degrade model performance.

In this study, we propose  a novel two-step \textbf{Mo}mentum encoder-utilized \textbf{SSDA} framework, \textbf{MoSSDA}, for multivariate time-series classification.
We evaluated our method using benchmark real-world, four different multivariate time series datasets \cite{kwapisz2011activityWISDM, stisen2015HHAR, anguita2013UCIHAR, wagner10ptbxl}, and two different univariate time series datasets \cite{lessmeier2016MFD,goldberger2000EEG}, and achieved state-of-the-art performances.

The key contributions of this study are summarized as follows:
\begin{itemize}

\item We propose a novel SSDA strategy that emphasizes the extraction of robust, domain-invariant, and inter- and intra-domain class-discriminative representations, using three different modules and a decoupled two-step training process.
\item A domain-invariant encoder that mitigates domain shifts by utilizing both unlabeled and labeled data, was employed using the maximum mean discrepancy (MMD) loss to learn domain-invariant features.
\item In the positive contrastive module, inter- and intra-domain representations are learned through supervised contrastive loss with a mixup. Mixup allows MoSSDA to efficiently utilize limited target domain data as rich representations without the data augmentation process commonly used in prior SSDA methods.
\item A momentum encoder was employed to enhance feature consistency and discriminability by preserving semantically meaningful representations across iterations.
\item Through extensive experiments on time-series classification, we demonstrated that the proposed method achieves state-of-the-art performance, significantly outperforming existing widely used SSDA approaches.

\end{itemize}

\section{Related Work} \label{sec:relatedwork}

\subsection{Time Series Domain Adaptation} \label{subsec:TSDA}
The objective of time-series domain adaptation is to learn feature representations that are domain-invariant (ensuring robustness across domains) and class-discriminative (ensuring classification effectiveness) \cite{ott2022domainLabelShift, shi2022deepSurvey,chen2024CADT,liu2021AdvSKM}. To meet this requirement, specialized methods have been developed to preserve the unique temporal characteristics of the data.
Most approaches have focused on the alignment of feature distributions across domains. A prominent strategy involves minimizing statistical distance metrics, such as MMD in a Reproducing Kernel Hilbert Space (RKHS) \cite{ott2022domainLabelShift}. AdvSKM \cite{liu2021AdvSKM} was proposed with a novel spectral kernel to enhance the MMD metric for more accurate discrepancy measurements. In another study, MMD was combined with feature transformation techniques, such as correlation alignment, to map source features more closely to the target distribution \cite{he2023domain}. Other significant research efforts have utilized adversarial training. Adversarial-training-based methods comprise a domain discriminator that distinguishes features from the source and target domains, whereas the feature extractor is trained to generate domain-agnostic features \cite{wilson2020CoDATS}. For instance, DAF \cite{jin2022domain} incorporates a domain discriminator with shared attention modules for time-series forecasting. Recent methods, such as CADT \cite{chen2024CADT}, focus on disentangling domain-invariant features from domain-specific one, and use custom contrastive learning objectives to address the instabilities common in adversarial architectures. Contrastive learning has recently emerged as an effective technique for domain alignment in time-series studies. CoTMix \cite{eldele2023CoTMix} is an example that exclusively utilizes contrastive objectives to mitigate distribution shifts. DACAD \cite{darban2024dacad} integrates contrastive learning with the UDA for anomaly detection by incorporating supervised and self-supervised contrastive losses into the source and target domains respectively.

\subsection{Semi-Supervised Domain Adaptation} \label{subsec:SSDA}
In the SSDA framework, a limited number of labeled samples are available in the target domain. These samples can be utilized in conjunction with a substantial unlabeled data corpus to significantly enhance performance \cite{yoon2022sampletosample,cheng2014SSDAmanifold}. A prevalent technique in SSDA involves the application of consistency regularization, which is frequently accompanied by data augmentation. AdaMatch \cite{berthelot2021adamatch} enforces consistency between the predictions on weakly and strongly augmented versions of the target samples to align the distributions. Similarly, DARK \cite{kim2022DARK} employs cross-view consistency regularization to distill domain-specific knowledge. Another study demonstrated a combining self-supervised pretraining with consistency regularization can yield strong results without explicit domain alignment \cite{mishra2021PAC}. Pseudo-labeling is another prevalent SSDA method. This method leverages the model's high-confidence predictions of unlabeled target data as "pseudo-labels" to expand the training set. DECOTA \cite{yang2021deepCoTraining} employs a co-training framework that utilizes pseudo-labels to decompose the SSDA tasks. However, naïve pseudo-labeling can reinforce confirmation bias. Methods, such as UniSSDA \cite{zhang2024uniSSDA}, have proposed prior-guided refinement strategies to mitigate this issue, particularly in challenging settings with private classes. Other approaches have focused on adversarial training and clustering. For instance, a minimax entropy approach was proposed to adversarially optimize a few-shot model \cite{saito2019minimaxSSDA}, whereas CDAC \cite{li2021CDAC} utilizes adversarial adaptive clustering loss to align inter- and intra-domain distributions. Various prevailing methodologies depend on sophisticated data augmentation strategies \cite{kim2022DARK,berthelot2021adamatch,li2021ecacl} or intricate end-to-end adversarial training \cite{ganin2015adversarialUDA,long2018CDAN,shu2018dirt}.

\section{Methodology} \label{sec:methodology}
 
\subsection{Problem Formulation} \label{subsec:problem}
In the SSDA setting for multivariate time-series classification, source and target domain datasets were provided. The source domain dataset is fully labeled, while the target domain dataset contains only a few labeled samples, with the remainder unlabeled. The source domain dataset is expressed as follows: $\mathcal{D}_{\text{src}} = \{ (X_i^{\text{src}}, y_i^{\text{src}}) \}_{i=1}^{N_s}$, where $N_s$ is the number of source samples. Each sample, $X_i^{\text{src}} \in \mathbb{R}^{D \times T}$, is a multivariate time-series instance with $D$ variables (channels) and $T$ time steps, and $y_i^{\text{src}} \in \mathcal{Y}$ is its corresponding class label. The target domain dataset, $\mathcal{D}_{\text{trg}}$, consisted of two distinct subsets. A small labeled set, $\mathcal{D}_{\text{trg}}^{\ell} = \{ (X_j^{\text{trg}}, y_j^{\text{trg}}) \}_{j=1}^{N_t^\ell}$, where $N_t^\ell$ is the number of labeled target samples, and a large unlabeled set, $\mathcal{D}_{\text{trg}}^{u} = \{ X_k^{\text{trg}} \}_{k=1}^{N_t^u}$, where $N_t^u$ is the number of unlabeled target samples. Naturally, the number of labeled target data is smaller than that of unlabeled ones, i.e., $N_t^\ell \ll N_t^u$. We denote the combined set of all available labeled data as $\mathcal{D}^{\ell} = \mathcal{D}_{\text{src}} \cup \mathcal{D}_{\text{trg}}^{\ell}$.
We assume that the source and target domains share the same label space $\mathcal{Y}$ but their data distributions $P(X^{\text{src}})$ and $P(X^{\text{trg}})$ can be different. The main objective of our framework is to minimize the domain discrepancy between the source and target domains, and to learn a task-specific classifier using $\mathcal{D}_{\text{src}}$ and $\mathcal{D}_{\text{trg}}$ to accurately predict labels on test data from the target domain.

\subsection{Method Description} \label{subsec:methodDescription}

\begin{figure}[t]
    \centering
    \includegraphics[height=0.88\linewidth]{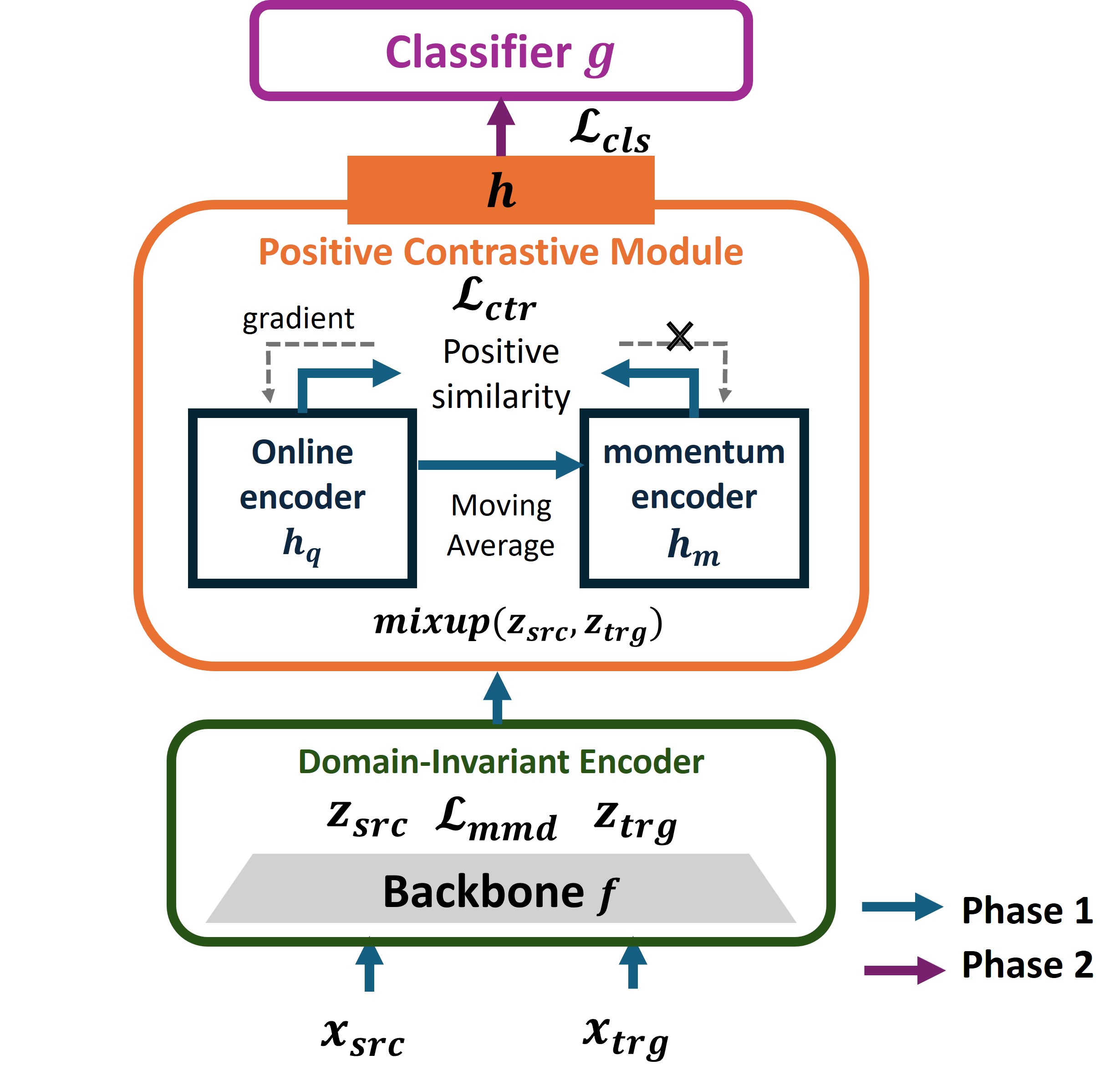}
    \caption{{\bf Overview of MoSSDA framework.} In the decoupled two-step framework, the first step involves updating the components (colored in navy): The following part is used to update the classifier (colored in purple). The weights in the other steps were not updated for each step.}
    \label{fig:figure1}
\end{figure}

The framework of the proposed {\bf MoSSDA} for multivariate time-series semi-supervised domain adaptation is shown in Figure \ref{fig:figure1}. In the first stage, the framework utilizes the full spectrum of available data, labeled source domain data $\mathcal{D}_{\text{src}}$, labeled target domain data $\mathcal{D}_{\text{trg}}^{\ell}$, and unlabeled target domain data $\mathcal{D}_{\text{trg}}^{u}$, to learn feature representations.
The learned feature representations are simultaneously domain-invariant and class-discriminative, promoting both intra- and inter-domain class separability by employing {\bf a Domain-Invariant Encoder} and {\bf Positive Contrastive Module}. 
The second stage trains a high-performance {\bf Classification Module} on the learned feature representation using only the reliable labeled data $\mathcal{D}^{\ell}$. This separation prevents conflicting optimization objectives and enhances training stability.

\subsection{Domain-Invariant Encoder} \label{subsec:domaininvairantEncoder}

The first component of a framework is the domain-invariant encoder, which is designed to learn feature representations that are robust to distributional shifts between the source and target domains. We denote this feature extractor as a network $f$, parameterized by $\theta_f$. The encoder maps an input time-series $X$ to a latent representation $Z = f(X)$. Let $Z_{\text{src}} = f(X_{\text{src}})$ and $Z_{\text{trg}} = f(X_{\text{trg}}) = \{Z_{\text{trg}}^{\ell},Z_{\text{trg}}^{u} \}$ be the sets of feature representations for the source and target domains, respectively.

To obtain domain invariance, we employed MMD loss, which is a widely used metric for comparing distributions in RKHS. The squared MMD between the source and target feature distributions was used as the MMD loss, which can be computed over batches as follows:

\begin{equation}
\label{eq:mmd_loss}
\mathcal{L}_{\text{mmd}} = \left\| \mathbb{E}_{z_s \sim Z_{\text{src}}}[ \phi(z_s) ] - \mathbb{E}_{z_t \sim Z_{\text{trg}}}[ \phi(z_t) ] \right\|_{\mathcal{H}}^2
\end{equation}
where $z_s \in Z_{\text{src}}$, $z_t \in Z_{\text{trg}}$, and $\phi(\cdot)$ is a mapping to the RKHS $\mathcal{H}$. The empirical estimate of this loss, using the kernel trick $k(x, x') = \langle \phi(x), \phi(x') \rangle$
In this work, we primarily used a linear kernel, $k(x, x') = x^\top x'$, for its simplicity and computational efficiency. However, our framework is flexible, allowing for the substitution of other kernels (e.g., RBF) as required by the specific dataset characteristics.

\subsection{Positive Contrastive Module} \label{subsec:positivecontrastiveModule}
The second key objective of the framework is to learn feature representations that are class-discriminative both within each domain (intra-domain) and across them (inter-domain). This can be obtained by leveraging the labeled data from both domains, $\mathcal{D}^{\ell}$. A significant challenge arises from scarcity of labeled target data, which can lead to the feature space to be highly biased towards the source domain. To compensate for this bias and promote robust feature learning, we designed a strategy that combines mixup \cite{zhang2017mixup} with a supervised contrastive loss \cite{khosla2020supervisedCL,grill2020BYOL}.

We created a positive counterpart of $z_i$ by linearly interpolating its feature representation with that of another sample ($z_j$) from the \textbf{same class}, which may belong to the source or target domain, as follows:
\begin{equation}
\label{eq:mixup}
z_{\text{mix}} = \lambda z_i + (1-\lambda)z_j \quad 
\end{equation}
where $ y_i=y_j, \lambda \sim \text{Beta}(\alpha, \alpha)$
This process enables the model to learn smoother decision boundaries and creates a more diverse set of positive examples for the subsequent contrastive learning step. The set of latent representations used for this module is denoted $\mathcal{Z}^\ell = \{Z_{\text{src}}, Z_{\text{trg}}^{\ell}, Z_{\text{mix}}\}$, thus comprises original labeled features from both domains and their mixed-up counterparts.
With the enriched labeled set ${Z}^\ell$, we train the encoder using a supervised contrastive loss \cite{khosla2020supervisedCL,grill2020BYOL,chen2021noNegative}. This loss function encourages feature representations of samples from the same classes (positive) should be closer together in the embedding space. For a training batch, the loss can be defined as:
\begin{equation}
\label{eq:supcon}
\mathcal{L}_{ctr}=\mathbb{E}\left[-\log{\frac{\sum_{j\neq i} I\left(y_i=y_j\right)\exp{\left(\mathrm{sim}\left(z_i,z_j\right)/\tau\right)}}{\sum_{k}\exp{\left(\mathrm{sim}\left(z_i,z_k\right)/\tau\right)}}}\right]
\end{equation}
where $\tau$ is a temperature hyperparameter, and $\text{sim}(u, v) = u^\top v / (\|u\| \|v\|)$ denotes the cosine similarity. 
The objective of this loss function is to leverage class labels to capture the embedding space and encourage the samples to belong to the same class—regardless of their domain of origin—to formed compact and well-separated clusters. Specifically, positive pairs are constructed using a combination of samples from the source domain, target domain, and their interpolated mixtures via a mixup. This design integrates both intra- and inter-domain positive pairs into the learning objective, thereby encouraging the acquisition of class-discriminative and domain-robust representations.

To ensure that the feature representations used for contrastive learning were stable and consistent, we employed the momentum encoder pioneered by MoCo \cite{he2020MoCo}. Thus, the positive contrastive module comprises two encoders.

\begin{itemize}
    \item \textbf{An online encoder}: $h_q$ parameterized by weights $\theta_q$. These weights are actively updated via backpropagation from the learning objectives.
    \item \textbf{A momentum encoder}: $h_m$ parameterized by weights $\theta_m$. These weights were \textbf{not} updated via backpropagation.
\end{itemize}

Instead of direct gradient updates, the momentum encoder's weights, $\theta_m$, are updated as an \textit{exponentially moving average} (EMA) of the online encoder's weights, $\theta_q$. After each training step, an update was performed as follows:
\begin{equation}
\label{eq:momentum_update}
\theta_m \leftarrow m \cdot \theta_m + (1-m) \cdot \theta_q
\end{equation}
where $m \in [0, 1)$ is the momentum coefficient, a hyperparameter that controls the speed of the update. $\theta_q$ represents the weight of the online encoder after the gradient update in the current training step. $\theta_m$ represents the weights of the momentum encoder, which are being updated. The momentum coefficient $m$ is typically set to a large value, such as 0.999.
This is highly beneficial for contrastive learning because it provides a stable and consistent feature representation. This prevents instability in the learning process, which can occur if the feature keys in the contrastive dictionary change rapidly at every gradient step. Consequently, the model can effectively learn robust time-series features without requiring a large labeled training batch.

\subsection{Classification Module} \label{subsec:classificationModule}
In the first stage, a domain-invariant encoder is trained with a positive contrastive learning module, thereby producing feature representations that are both domain-invariant and class-discriminative across and within domains. 
In the second stage, we train a high-performance classifier using this enhanced labeled set. All the modules learned in the first stage were frozen, and the classifier was trained solely on the optimized features extracted from the frozen encoder.

The classification loss employed is the standard cross-entropy loss, defined as:
\begin{equation}
\label{eq:cross_entropy loss}
\mathcal{L}_{\mathrm{ce}} = - \sum_{i=1}^{N} y_i \log \left( \widehat{y}_i \right) 
\end{equation}
where $y_i$ denotes the ground truth label and $\widehat{y}_i$ the predicted probability of the $i$-th sample. The final classification loss can be defined as:

\begin{equation}
\label{eq:classification loss}
\mathcal{L}_{\mathrm{cls}} = \mathcal{L}_{\mathrm{ce}}^{s} + \mathcal{L}_{\mathrm{ce}}^{t}
\end{equation}
where $\mathcal{L}_{\mathrm{ce}}^{s}$ and $\mathcal{L}_{\mathrm{ce}}^{t}$ depict the cross-entropy losses on the source and target domains, respectively. This loss function encourages the classifier to learn class-discriminative decision boundaries, thereby improving its ability to correctly classify samples from both source and target domains based on their respective labels.

\subsection{Overall Loss} \label{subsec:overallLoss}
The overall training objective consisting of the MMD loss, positive contrastive loss, and classification loss can be formulated as follows:
\begin{equation}
\label{eq:overall loss}
\mathcal{L}_{\mathrm{tot}} = \lambda_{\text{mmd}}*\mathcal{L}_{\text{mmd}} + \lambda_{\text{ctr}}*\mathcal{L}_{ctr} + \mathcal{L}_{\mathrm{cls}}
\end{equation}
where $\lambda_{\text{mmd}}$ and $\lambda_{\text{ctr}}$ are hyperparameters that balance the contribution of each loss term.
Specifically, the overall loss for the first stage is a weighted sum of the MMD loss and the supervised contrastive loss, as follows:
\begin{equation}
\label{eq:stage1_loss}
\mathcal{L}_{\text{stage1}} = \lambda_{\text{mmd}}* \mathcal{L}_{\text{mmd}} + \lambda_{\text{ctr}}* \mathcal{L}_{\text{ctr}}
\end{equation}
The loss in the second stage minimizes the standard cross-entropy loss over all available reliably labeled data.
\begin{equation}
\label{eq:stage2_loss}
\mathcal{L}_{\text{stage2}} = \mathcal{L}_{\text{ce}}(\mathcal{D}^\ell) = \mathcal{L}_{\text{ce}}^{s} + \mathcal{L}_{\text{ce}}^{t}
\end{equation}

Simultaneous training of both stages may cause the first stage to create a simple latent representation that can be easily classified in the second stage, potentially leading to an overfitting. Hence, we train the first stage using Eq. \ref{eq:stage1_loss} to construct a robust, domain-agnostic feature space and, then train the second stage with Eq. \ref{eq:stage2_loss} to learn the optimal decision boundaries within space.

\section{Experiments} \label{sec:Experiments}

\begin{table*}[p]
\centering
\resizebox{\textwidth}{!}{%
\begin{tabular}{cc|ccccccc||ccccccc}
\hline\hline
 &  & \multicolumn{7}{c||}{\textbf{\rule{0pt}{12pt}Averaged test accuracy}} & \multicolumn{7}{c}{\textbf{Averaged test f1-score}} \\
\rule{0pt}{12pt}dataset & unlab. ratio ($u$)& AdaMatch & CDAC & DST & PAC & UniSSDA & \multicolumn{1}{c|}{CLDA} & MoSSDA & AdaMatch & CDAC & DST & PAC & UniSSDA & \multicolumn{1}{c|}{CLDA} & MoSSDA \\ \hline
\multirow{3}{*}{EEG} & 0.7 & 0.5107 & 0.1282 & 0.5045 & \underline{ 0.6144} & 0.4527 & \multicolumn{1}{c|}{0.3999} & \textbf{0.7910} & 0.3837 & 0.0478 & 0.3885 & \underline{ 0.5029} & 0.3455 & \multicolumn{1}{c|}{0.3257} & \textbf{0.6862} \\
 & 0.9 & 0.4767 & 0.1282 & 0.4609 & \underline{ 0.4835} & 0.4160 & \multicolumn{1}{c|}{0.3184} & \textbf{0.7553} & \underline{ 0.3816} & 0.0479 & 0.3776 & 0.3545 & 0.3223 & \multicolumn{1}{c|}{0.2603} & \textbf{0.6552} \\
 & 0.95 & \underline{ 0.4684} & 0.1286 & 0.4558 & 0.4503 & 0.4046 & \multicolumn{1}{c|}{0.3147} & \textbf{0.7328} & 0.3806 & 0.0480 & \underline{ 0.3806} & 0.3064 & 0.3121 & \multicolumn{1}{c|}{0.2616} & \textbf{0.6244} \\ \hline
\multirow{3}{*}{HAR} & 0.7 & \underline{ 0.6097} & 0.1524 & 0.5931 & 0.4773 & 0.5465 & \multicolumn{1}{c|}{0.2639} & \textbf{0.9594} & \underline{ 0.5284} & 0.0496 & 0.5056 & 0.3972 & 0.4319 & \multicolumn{1}{c|}{0.1884} & \textbf{0.9606} \\
 & 0.9 & \underline{ 0.6001} & 0.1479 & 0.5880 & 0.2777 & 0.5191 & \multicolumn{1}{c|}{0.2665} & \textbf{0.9376} & \underline{ 0.5333} & 0.0461 & 0.4946 & 0.1682 & 0.3862 & \multicolumn{1}{c|}{0.1881} & \textbf{0.9392} \\
 & 0.95 & \underline{ 0.6071} & 0.1501 & 0.5871 & 0.2889 & 0.5221 & \multicolumn{1}{c|}{0.2660} & \textbf{0.8970} & \underline{ 0.5332} & 0.0484 & 0.4912 & 0.1656 & 0.3956 & \multicolumn{1}{c|}{0.1902} & \textbf{0.8948} \\ \hline
\multirow{3}{*}{HHAR} & 0.7 & 0.5130 & 0.1651 & 0.5150 & \underline{ 0.5449} & 0.5107 & \multicolumn{1}{c|}{0.4897} & \textbf{0.9693} & 0.4949 & 0.0532 & \underline{ 0.4987} & 0.4954 & 0.4914 & \multicolumn{1}{c|}{0.4536} & \textbf{0.9698} \\
 & 0.9 & 0.5093 & 0.1654 & \underline{ 0.5159} & 0.4228 & 0.4987 & \multicolumn{1}{c|}{0.4632} & \textbf{0.9563} & 0.4935 & 0.0535 & \underline{ 0.5071} & 0.3587 & 0.4830 & \multicolumn{1}{c|}{0.4293} & \textbf{0.9567} \\
 & 0.95 & \underline{ 0.5096} & 0.1651 & 0.5043 & 0.2822 & 0.4998 & \multicolumn{1}{c|}{0.4615} & \textbf{0.9430} & \underline{ 0.4915} & 0.0532 & 0.4900 & 0.1817 & 0.4816 & \multicolumn{1}{c|}{0.4238} & \textbf{0.9433} \\ \hline
\multirow{3}{*}{MFD} & 0.7 & 0.6990 & 0.4550 & 0.5805 & \underline{ 0.8036} & 0.6844 & \multicolumn{1}{c|}{0.6910} & \textbf{0.9726} & 0.6868 & 0.2085 & 0.6216 & \underline{ 0.7937} & 0.6739 & \multicolumn{1}{c|}{0.7096} & \textbf{0.9571} \\
 & 0.9 & 0.6932 & 0.4550 & \underline{ 0.6949} & 0.3772 & 0.6801 & \multicolumn{1}{c|}{0.6788} & \textbf{0.9339} & 0.6811 & 0.2085 & 0.6803 & 0.2751 & 0.6709 & \multicolumn{1}{c|}{\underline{ 0.6898}} & \textbf{0.9065} \\
 & 0.95 & \underline{ 0.6907} & 0.4550 & 0.5622 & 0.6584 & 0.6801 & \multicolumn{1}{c|}{0.6764} & \textbf{0.9519} & 0.6784 & 0.2085 & 0.5849 & 0.6444 & 0.6720 & \multicolumn{1}{c|}{\underline{ 0.6858}} & \textbf{0.9096} \\ \hline
\multirow{3}{*}{PTBXL} & 0.7 & 0.4816 & 0.2197 & 0.4521 & \underline{ 0.4869} & 0.4867 & \multicolumn{1}{c|}{0.4835} & \textbf{0.7361} & 0.2720 & 0.0934 & 0.2585 & \underline{ 0.3939} & 0.2157 & \multicolumn{1}{c|}{0.2254} & \textbf{0.6179} \\
 & 0.9 & 0.4402 & 0.2236 & 0.4044 & \underline{ 0.5359} & 0.4350 & \multicolumn{1}{c|}{0.4380} & \textbf{0.7213} & \underline{ 0.2700} & 0.0953 & 0.2620 & 0.2573 & 0.2162 & \multicolumn{1}{c|}{0.2667} & \textbf{0.5880} \\
 & 0.95 & 0.4302 & 0.2239 & 0.4013 & \underline{ 0.5167} & 0.4064 & \multicolumn{1}{c|}{0.3791} & \textbf{0.7014} & \underline{ 0.2831} & 0.0958 & 0.2715 & 0.2144 & 0.2098 & \multicolumn{1}{c|}{0.2726} & \textbf{0.5701} \\ \hline
\multirow{3}{*}{WISDM} & 0.7 & 0.3358 & 0.0897 & \underline{ 0.5356} & 0.5035 & 0.3486 & \multicolumn{1}{c|}{0.1352} & \textbf{0.7838} & 0.0996 & 0.0450 & \underline{ 0.3503} & 0.2018 & 0.1817 & \multicolumn{1}{c|}{0.1006} & \textbf{0.7176} \\
 & 0.9 & 0.2419 & 0.0883 & \underline{ 0.5091} & 0.3548 & 0.3697 & \multicolumn{1}{c|}{0.1305} & \textbf{0.7403} & 0.0657 & 0.0399 & \underline{ 0.3599} & 0.1222 & 0.1831 & \multicolumn{1}{c|}{0.0998} & \textbf{0.6709} \\
 & 0.95 & 0.2079 & 0.0966 & \underline{ 0.4991} & 0.3389 & 0.3766 & \multicolumn{1}{c|}{0.1353} & \textbf{0.6732} & 0.0703 & 0.0499 & \underline{ 0.3584} & 0.1225 & 0.2093 & \multicolumn{1}{c|}{0.1021} & \textbf{0.6084} \\ \hline\hline
\end{tabular}%
}
\caption{{\bf Comparison with SSDA methods}: Averaged target domain test accuracy and f1-score across domain pairs for each datasets with RESNET18 backbone. The best performance is in bold and the second best is underlined.}
\label{baselinesRESNET18}
\end{table*}
\begin{table*}[]
\centering
\resizebox{\textwidth}{!}{%
\begin{tabular}{cc|ccccccc||ccccccc}
\hline\hline
 &  & \multicolumn{7}{c||}{\textbf{\rule{0pt}{12pt}Averaged test accuracy}} & \multicolumn{7}{c}{\textbf{Averaged test f1-score}} \\
\rule{0pt}{12pt}dataset & unlab. ratio ($u$)& AdaMatch & CDAC & DST & PAC & UniSSDA & \multicolumn{1}{c|}{CLDA} & MoSSDA & AdaMatch & CDAC & DST & PAC & UniSSDA & \multicolumn{1}{c|}{CLDA} & MoSSDA \\ \hline
\multirow{3}{*}{EEG}                       & 0.7                & 0.4864   & 0.2217 & 0.4983       & \underline{ 0.6469} & 0.4499  & \multicolumn{1}{c|}{0.3676} & \textbf{0.8369} & 0.3816   & 0.0768 & 0.4017       & \underline{ 0.5353} & 0.3382  & \multicolumn{1}{c|}{0.2500} & \textbf{0.7555} \\
                                           & 0.9                & 0.4638   & 0.2237 & 0.4572       & \underline{ 0.5646} & 0.4028  & \multicolumn{1}{c|}{0.3268} & \textbf{0.8057} & 0.3755   & 0.0779 & 0.3865       & \underline{ 0.4338} & 0.3147  & \multicolumn{1}{c|}{0.2296} & \textbf{0.7245} \\
                                           & 0.95               & 0.4576   & 0.2219 & 0.4522       & \underline{ 0.4957} & 0.3974  & \multicolumn{1}{c|}{0.3111} & \textbf{0.7813} & 0.3718   & 0.0770 & \underline{ 0.3879} & 0.3501       & 0.3121  & \multicolumn{1}{c|}{0.2115} & \textbf{0.6991} \\ \hline
\multirow{3}{*}{HAR}                       & 0.7                & 0.5020   & 0.1513 & 0.5142       & \underline{ 0.6838} & 0.4372  & \multicolumn{1}{c|}{0.4525} & \textbf{0.9708} & 0.3873   & 0.0430 & 0.4071       & \underline{ 0.6273} & 0.3238  & \multicolumn{1}{c|}{0.3227} & \textbf{0.9704} \\
                                           & 0.9                & 0.5071   & 0.1513 & 0.5008       & \underline{ 0.5076} & 0.4240  & \multicolumn{1}{c|}{0.4539} & \textbf{0.9647} & 0.3936   & 0.0430 & 0.3999       & \underline{ 0.4113} & 0.3096  & \multicolumn{1}{c|}{0.3140} & \textbf{0.9645} \\
                                           & 0.95               & 0.4947   & 0.1513 & 0.4993       & \underline{ 0.5567} & 0.4155  & \multicolumn{1}{c|}{0.4611} & \textbf{0.9279} & 0.3696   & 0.0430 & 0.3958       & \underline{ 0.4666} & 0.3022  & \multicolumn{1}{c|}{0.3225} & \textbf{0.9161} \\ \hline
\multirow{3}{*}{HHAR}                      & 0.7                & 0.4380   & 0.1383 & 0.4393       & \underline{ 0.5897} & 0.4407  & \multicolumn{1}{c|}{0.4195} & \textbf{0.9784} & 0.3897   & 0.0414 & 0.3986       & \underline{ 0.5504} & 0.3945  & \multicolumn{1}{c|}{0.3813} & \textbf{0.9788} \\
                                           & 0.9                & 0.4416   & 0.1412 & 0.4525       & \underline{ 0.5330} & 0.4541  & \multicolumn{1}{c|}{0.4370} & \textbf{0.9595} & 0.3973   & 0.0423 & 0.4131       & \underline{ 0.4740} & 0.4118  & \multicolumn{1}{c|}{0.4022} & \textbf{0.9604} \\
                                           & 0.95               & 0.4415   & 0.1421 & \underline{ 0.4499} & 0.4197       & 0.4457  & \multicolumn{1}{c|}{0.4325} & \textbf{0.9480} & 0.3957   & 0.0425 & \underline{ 0.4113} & 0.3492       & 0.4011  & \multicolumn{1}{c|}{0.4013} & \textbf{0.9494} \\ \hline
\multirow{3}{*}{MFD}                       & 0.7                & 0.5544   & 0.4539 & 0.5775       & \underline{ 0.7957} & 0.5707  & \multicolumn{1}{c|}{0.5577} & \textbf{0.9832} & 0.6036   & 0.2081 & 0.6183       & \underline{ 0.7843} & 0.6153  & \multicolumn{1}{c|}{0.4977} & \textbf{0.9793} \\
                                           & 0.9                & 0.5501   & 0.4539 & \underline{ 0.6953} & 0.6882       & 0.5538  & \multicolumn{1}{c|}{0.5679} & \textbf{0.9777} & 0.6008   & 0.2081 & \underline{ 0.6824} & 0.6784       & 0.6037  & \multicolumn{1}{c|}{0.5047} & \textbf{0.9759} \\
                                           & 0.95               & 0.5547   & 0.4539 & 0.5676       & \underline{ 0.7331} & 0.5596  & \multicolumn{1}{c|}{0.5721} & \textbf{0.9798} & 0.6048   & 0.2081 & 0.5904       & \underline{ 0.7529} & 0.6079  & \multicolumn{1}{c|}{0.5090} & \textbf{0.9736} \\ \hline
\multirow{3}{*}{PTBXL}                     & 0.7                & 0.4520   & 0.1732 & 0.5082       & \underline{ 0.5841} & 0.5018  & \multicolumn{1}{c|}{0.5690} & \textbf{0.7294} & 0.1910   & 0.0632 & 0.2132       & \underline{ 0.4432} & 0.1916  & \multicolumn{1}{c|}{0.2748} & \textbf{0.5962} \\
                                           & 0.9                & 0.4454   & 0.1735 & 0.5061       & \underline{ 0.5944} & 0.4914  & \multicolumn{1}{c|}{0.4868} & \textbf{0.7209} & 0.2043   & 0.0642 & 0.2276       & \underline{ 0.4040} & 0.1885  & \multicolumn{1}{c|}{0.2325} & \textbf{0.5719} \\
                                           & 0.95               & 0.4459   & 0.1737 & \underline{ 0.5025} & 0.4857       & 0.4061  & \multicolumn{1}{c|}{0.4754} & \textbf{0.7005} & 0.2166   & 0.0650 & 0.2303       & \underline{ 0.2917} & 0.1715  & \multicolumn{1}{c|}{0.2209} & \textbf{0.5632} \\ \hline
\multirow{3}{*}{WISDM}                     & 0.7                & 0.3422   & 0.1344 & 0.3306       & \underline{ 0.5591} & 0.3348  & \multicolumn{1}{c|}{0.2444} & \textbf{0.8357} & 0.0923   & 0.0349 & 0.2195       & \underline{ 0.3202} & 0.2221  & \multicolumn{1}{c|}{0.1577} & \textbf{0.7987} \\
                                           & 0.9                & 0.3020   & 0.1344 & 0.3365       & \underline{ 0.5964} & 0.3274  & \multicolumn{1}{c|}{0.2262} & \textbf{0.7323} & 0.0795   & 0.0349 & 0.2399       & \underline{ 0.3886} & 0.2172  & \multicolumn{1}{c|}{0.1476} & \textbf{0.6536} \\
                                           & 0.95               & 0.2795   & 0.1344 & 0.3074       & \underline{ 0.4632} & 0.3075  & \multicolumn{1}{c|}{0.2070} & \textbf{0.6459} & 0.0988   & 0.0349 & 0.2585       & \underline{ 0.3017} & 0.2145  & \multicolumn{1}{c|}{0.1532} & \textbf{0.5745} \\ \hline
\end{tabular}%
}
\caption{{\bf Comparison with SSDA methods}: Averaged target domain test accuracy and f1-score across domain pairs for each datasets with CNN backbone. The best performance is in bold and the second best is underlined.}
\label{baselinesCNN}
\end{table*}

\begin{table*}[]
\centering
\resizebox{\textwidth}{!}{%
\begin{tabular}{cc|ccccccc||ccccccc}
\hline\hline
 &  & \multicolumn{7}{c||}{\textbf{\rule{0pt}{12pt}Averaged test accuracy}} & \multicolumn{7}{c}{\textbf{Averaged test f1-score}} \\
\rule{0pt}{12pt}dataset & unlab. ratio ($u$)& AdaMatch & CDAC & DST & PAC & UniSSDA & \multicolumn{1}{c|}{CLDA} & MoSSDA & AdaMatch & CDAC & DST & PAC & UniSSDA & \multicolumn{1}{c|}{CLDA} & MoSSDA \\ \hline
\multirow{3}{*}{EEG}                       & 0.7                & 0.3465       & 0.1295 & \underline{ 0.3509}    & 0.3489          & 0.2575       & \multicolumn{1}{c|}{0.2973}       & \textbf{0.4863}       & \underline{ 0.2558} & 0.0652 & 0.2536          & 0.1965 & 0.1660  & \multicolumn{1}{c|}{0.1946}       & \textbf{0.3739}       \\
                                           & 0.9                & 0.3089       & 0.1297 & \underline{ 0.3146}    & 0.3084          & 0.2468       & \multicolumn{1}{c|}{0.2700}       & \textbf{0.4803}       & \underline{ 0.2524} & 0.0652 & 0.2499          & 0.1718 & 0.1589  & \multicolumn{1}{c|}{0.1900}       & \textbf{0.3596}       \\
                                           & 0.95               & 0.3001       & 0.1291 & 0.3081          & \underline{ 0.3137}    & 0.2423       & \multicolumn{1}{c|}{0.2586}       & \textbf{0.4695}       & 0.2469       & 0.0649 & \underline{ 0.2483}    & 0.1525 & 0.1565  & \multicolumn{1}{c|}{0.1883}       & \textbf{0.3597}       \\ \hline
\multirow{3}{*}{HAR}                       & 0.7                & 0.6240       & 0.1476 & \underline{ 0.6287}    & 0.1846          & 0.6227       & \multicolumn{1}{c|}{0.6002}       & \textbf{0.9444}       & \underline{ 0.5498} & 0.0572 & 0.5458          & 0.0623 & 0.5295  & \multicolumn{1}{c|}{0.5489}       & \textbf{0.9395}       \\
                                           & 0.9                & \underline{ 0.6487} & 0.1487 & 0.6324          & 0.1727          & 0.6072       & \multicolumn{1}{c|}{0.6036}       & \textbf{0.8963}       & \underline{ 0.5677} & 0.0574 & 0.5568          & 0.0490 & 0.5259  & \multicolumn{1}{c|}{0.5501}       & \textbf{0.8917}       \\
                                           & 0.95               & \underline{ 0.6400} & 0.1488 & 0.6358          & 0.1683          & 0.6037       & \multicolumn{1}{c|}{0.5867}       & \textbf{0.8604}       & 0.5478       & 0.0574 & \underline{ 0.5605}    & 0.0479 & 0.5151  & \multicolumn{1}{c|}{0.5277}       & \textbf{0.8516}       \\ \hline
\multirow{3}{*}{HHAR}                      & 0.7                & 0.5598       & 0.1542 & 0.5728          & 0.2363          & 0.5740       & \multicolumn{1}{c|}{\underline{ 0.6137}} & \textbf{0.9089}       & 0.5384       & 0.0656 & 0.5487          & 0.1133 & 0.5486  & \multicolumn{1}{c|}{\underline{ 0.5747}} & \textbf{0.9072}       \\
                                           & 0.9                & 0.5574       & 0.1524 & 0.5697          & 0.2903          & 0.5668       & \multicolumn{1}{c|}{\underline{ 0.6042}} & \textbf{0.9013}       & 0.5379       & 0.0644 & 0.5494          & 0.1926 & 0.5442  & \multicolumn{1}{c|}{\underline{ 0.5688}} & \textbf{0.9020}       \\
                                           & 0.95               & 0.5529       & 0.1534 & 0.5688          & 0.3078          & 0.5709       & \multicolumn{1}{c|}{\underline{ 0.6061}} & \textbf{0.8456}       & 0.5346       & 0.0655 & 0.5495          & 0.1990 & 0.5487  & \multicolumn{1}{c|}{\underline{ 0.5703}} & \textbf{0.8471}       \\ \hline
\multirow{3}{*}{MFD}                       & 0.7                & 0.5617       & 0.4550 & \underline{ 0.5744}    & 0.0910          & 0.5625       & \multicolumn{1}{c|}{0.5404}       & \textbf{0.6062}       & 0.5725       & 0.2085 & \underline{ 0.6168}    & 0.0556 & 0.5720  & \multicolumn{1}{c|}{0.5556}       & \textbf{0.6225}       \\
                                           & 0.9                & 0.5599       & 0.4550 & \textbf{0.6970} & 0.0910          & 0.5433       & \multicolumn{1}{c|}{0.5149}       & \underline{ 0.5923} & 0.5733       & 0.2085 & \textbf{0.6824} & 0.0556 & 0.5621  & \multicolumn{1}{c|}{0.5177}       & \underline{ 0.6062} \\
                                           & 0.95               & 0.5575       & 0.4550 & \underline{ 0.5630}    & 0.0910          & 0.5436       & \multicolumn{1}{c|}{0.5103}       & \textbf{0.5828}       & 0.5711       & 0.2085 & \underline{ 0.5861}    & 0.0556 & 0.5618  & \multicolumn{1}{c|}{0.5106}       & \textbf{0.6052}       \\ \hline
\multirow{3}{*}{PTBXL}                     & 0.7                & \underline{ 0.4519} & 0.1683 & 0.4169          & 0.3264          & 0.4522       & \multicolumn{1}{c|}{0.4088}       & \textbf{0.4284}       & 0.1981       & 0.1177 & 0.1853          & 0.1209 & 0.1657  & \multicolumn{1}{c|}{\underline{ 0.2346}} & \textbf{0.2551}       \\
                                           & 0.9                & 0.3998       & 0.1680 & 0.3264          & 0.2316          & \underline{ 0.4236} & \multicolumn{1}{c|}{0.3277}       & \textbf{0.4389}       & 0.2144       & 0.1187 & 0.2004          & 0.0701 & 0.1603  & \multicolumn{1}{c|}{\underline{ 0.2329}} & \textbf{0.2459}       \\
                                           & 0.95               & 0.3670       & 0.1683 & 0.2971          & \textbf{0.4788} & 0.4191       & \multicolumn{1}{c|}{0.2861}       & \underline{ 0.4451} & 0.2203       & 0.1183 & 0.2032          & 0.1519 & 0.1638  & \multicolumn{1}{c|}{\underline{ 0.2241}} & \textbf{0.2299}       \\ \hline
\multirow{3}{*}{WISDM}                     & 0.7                & 0.3314       & 0.1132 & 0.3133          & 0.1808          & 0.2130       & \multicolumn{1}{c|}{\underline{ 0.3586}} & \textbf{0.8203}       & 0.1057       & 0.0435 & 0.1057          & 0.0485 & 0.0535  & \multicolumn{1}{c|}{\underline{ 0.3050}} & \textbf{0.7020}       \\
                                           & 0.9                & 0.3055       & 0.1184 & 0.2945          & 0.1274          & 0.1385       & \multicolumn{1}{c|}{\underline{ 0.3430}} & \textbf{0.7811}       & 0.1032       & 0.0471 & 0.1190          & 0.0392 & 0.0379  & \multicolumn{1}{c|}{\underline{ 0.2950}} & \textbf{0.6538}       \\
                                           & 0.95               & 0.2950       & 0.1184 & 0.2796          & 0.1320          & 0.1059       & \multicolumn{1}{c|}{\underline{ 0.3252}} & \textbf{0.7031}       & 0.1212       & 0.0467 & 0.1230          & 0.0371 & 0.0302  & \multicolumn{1}{c|}{\underline{ 0.2725}} & \textbf{0.5943}       \\ \hline
\end{tabular}%
}
\caption{{\bf Comparison with SSDA methods}: Averaged target domain test accuracy and f1-score across domain pairs for each datasets with TCN backbone. The best performance is in bold and the second best is underlined.}
\label{baselinesTCN}
\end{table*}

\subsection{Experimental Setup} \label{subsec:expSetup}

\subsubsection{Datasets.} \label{subsubsec:datasets}
The experiment encompassed six time-series datasets from diverse domains, four multivariate time-series datasets, and two univariate time-series datasets, namely, UCIHAR \cite{anguita2013UCIHAR}, HHAR \cite{stisen2015HHAR}, WISDM \cite{kwapisz2011activityWISDM} EEG \cite{goldberger2000EEG}, PTBXL \cite{wagner2020ptbxl,wagner10ptbxl}, and MFD \cite{lessmeier2016MFD}. These datasets are commonly used for time-series domain adaptation tasks, except for the PTBXL dataset. A detailed description of the datasets is provided in the Supplementary S.\ref{subsec:dataset}

\subsubsection{Backbones.} \label{subsubsec:backbones}
Similar to previous studies on time-series domain adaptation \cite{ragab2023adatime,sun2024caudits,chen2024CADT}, we employed ResNet18 \cite{he2016deepRESNET18,fawaz2020deepRESNET18}, a CNN \cite{eldele2021timeCNN1,eldele2022adastCNN2}, and a TCN \cite{bai2018empiricalTCN1,thill2020timeTCN2} as the backbone networks in our experiments.
A 1D-CNN utilizes three convolutional blocks, each comprising a 1D-convolutional layer, BatchNorm, ReLU activation, and MaxPooling. RESNET18 implements a 1D residual network with shortcut connections between successive convolutional layers to enable deeper architectures. A TCN employs causal dilated convolutions to prevent temporal information leakage while capturing long-range dependencies in time-series data.
\subsubsection{Adaptation Scenarios.} \label{subsubsec:scenarios}
For a fair comparison, we used the same setting for the benchmark datasets as in the prior work \cite{ragab2023adatime}, including the data-splits and adaptation scenarios. In the case of PTBXL, we employed all six combinations of domains. We used a consistent setting across all experiments.

\subsubsection{Benchmark Methods.} \label{subsubsec:baselines}
State-of-the-art SSDA methods, CDAC \cite{li2021CDAC}, PAC \cite{mishra2021PAC}, AdaMatch \cite{berthelot2021adamatch}, and UniSSDA \cite{zhang2024uniSSDA}, were employed for comparison. In addition, we employed DST \cite{chen2022DST} and contrastive learning-based SSDA (CLDA) \cite{singh2021clda}. All the benchmark methods employed augmentation techniques, implemented by adapting image-specific augmentations to time-series-specific augmentations. 
\subsubsection{Implementation.} \label{subsubsec:implementation}
Each minibatch of size $B$ of source and target domain samples is equal, while the target domain samples consist of unlabeled data and labeled data, as provided in the unlabeled ratios: $u\in \{0.7, 0.9, 0.95\}$  across the entire experiments. We set hyperparameter temperature $\tau = 0.5$, momentum coefficient $m=0.999$, $\alpha = 1$, and both $\lambda_{\text{mmd}}$ and $\lambda_{\text{ctr}}$ set to 0.5. All the experiments were performed using PyTorch with an NVIDIA RTX 6000 Ada Generation system. The implementation details are described in Supplementary S.\ref{subsec:implementation}.

\subsection{Experimental Results} \label{subsec:expResults}

\subsubsection{Performance Comparison.} \label{subsubsec:ComparisonBaselines}
The proposed method was evaluated using three widely adopted backbone architectures, and its performance was compared with those of six state-of-the-art domain adaptation benchmark methods. The evaluation metrics included the mean accuracy and F1-score under the domain adaptation scenarios. Table \ref{baselinesRESNET18} presents the results obtained using RESNET18 as the underlying framework across six time series datasets with three distinct unlabeled ratios. Tables \ref{baselinesCNN} and \ref{baselinesTCN} present a comparison of the utilization of the CNN and TCN backbones under identical experimental conditions. These results demonstrate the adaptability of the proposed method to a wide range of backbone architectures.

The proposed method (MoSSDA) exhibited consistent superiority over other benchmark methods in target domain classification, particularly with the RESNET18 and CNN backbones. PAC demonstrated a competitive performance following MoSSDA; however, it experienced a substantial decrease in F1-score under more challenging settings (unlabeled ratio = 0.95). In contrast, MoSSDA demonstrated consistent performance, even when managing class-imbalanced datasets, such as PTBXL and WISDM. With the TCN (Table 3), a prevalent approach for time-series data, MoSSDA combines the optimal overall performance, followed by CLDA, which utilizes contrastive loss. These findings emphasize the necessity of selecting an appropriate backbone for the dataset to attain optimal domain adaptation performance. Further experiments with MLP and RNN-based backbones are provided in the Supplementary \ref{sec:supp}.

\begin{table}[h]
\centering
\resizebox{\linewidth}{0.13\linewidth}{%
\begin{tabular}{c|cccccc|cc} 
\hline\hline
              & \multicolumn{7}{c}{\rule{0pt}{13pt}\textbf{Averaged rank of averaged test accuracy}}           \\
$u$ & AdaMatch & CDAC & DST  & PAC  & UniSSDA & \multicolumn{1}{c|}{CLDA} & \textbf{OURS} \\ \hline
0.7           & 3.94     & 6.94 & 3.56 & 3.50 & 4.11    & \multicolumn{1}{c|}{4.83} & 1.11 \\
0.9           & 3.67     & 6.89 & 3.17 & 4.11 & 4.28    & \multicolumn{1}{c|}{4.83} & 1.06 \\
0.95          & 3.47     & 6.88 & 3.53 & 3.88 & 4.41    & \multicolumn{1}{c|}{4.76} & 1.06 \\ 
\hline\hline
\end{tabular}%
}
\caption{Averaged rank on overall results with SSDA methods.}
\label{averageRank}
\end{table}

Table \ref{averageRank} summarizes the results from Tables \ref{baselinesRESNET18} to \ref{baselinesTCN} by referring to the average ranking of each method across all experimental settings.
A lower rank indicates better performance. MoSSDA achieved the best overall ranking across all unlabeled ratio conditions, consistently outperforming the existing benchmark methods. Other semi-supervised methods varied depending on the amount of labeled data; however, MoSSDA retained its robustness. In the context of the relatively generous condition (unlabeled ratio = 0.7), PAC demonstrated the second-best performance, which can be attributed to its utilization of pretraining and consistency regularization. However, under more challenging conditions (unlabeled ratio = 0.9 and 0.95), DST and AdaMatch emerged as the next best performers. MoSSDA outperformed other methods that relied on pseudo-labeling or augmentation, because it leverages label information more effectively.

\subsubsection{Visulaization with t-SNE.} \label{subsubsec:visualization}

\begin{figure}[h]
\begin{minipage}[c][0.7\linewidth][c]{0.95\linewidth}
    \begin{subfigure}[b]{0.32\linewidth}
        \centering
        \includegraphics[width=\linewidth]{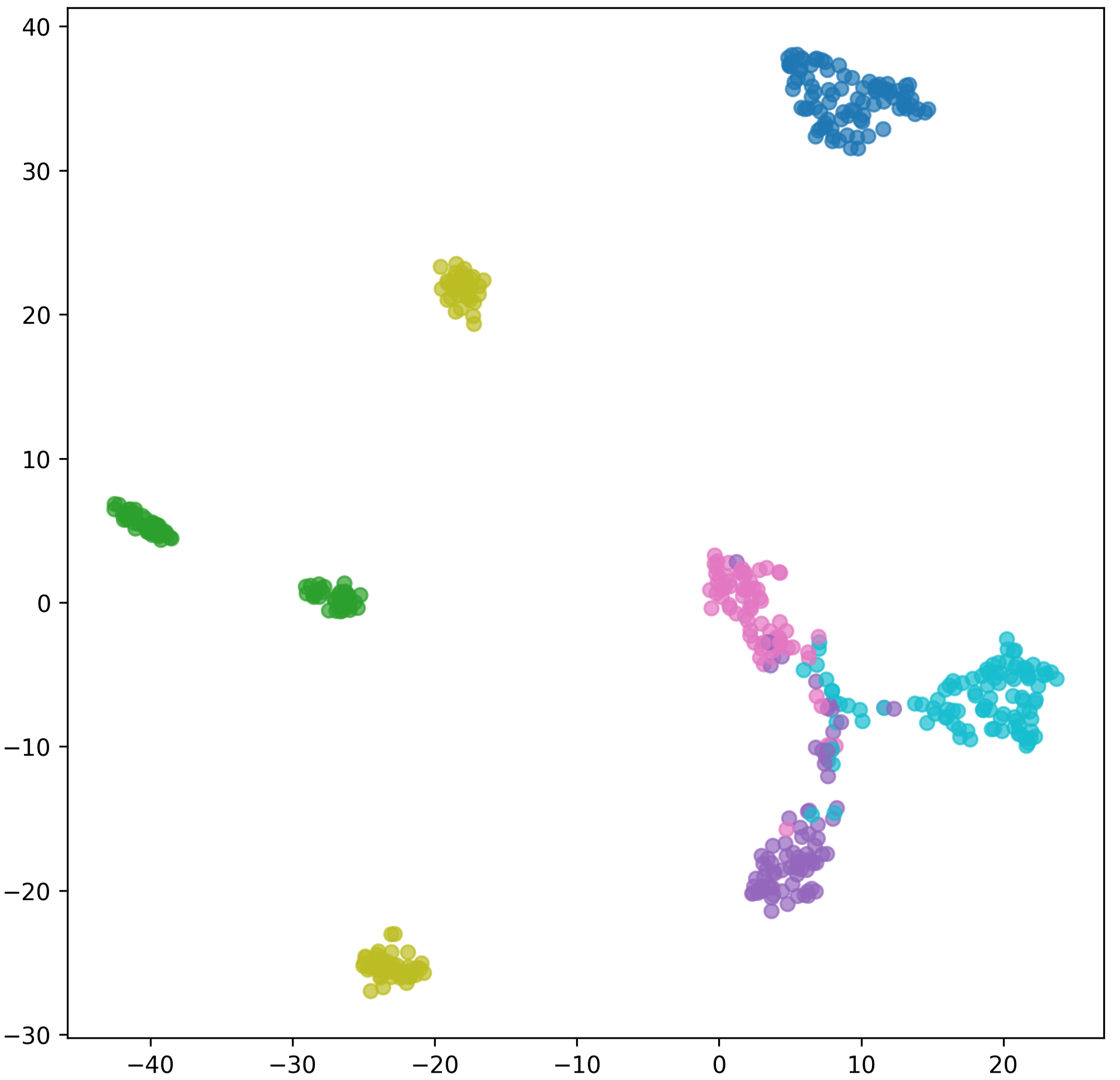}
        \caption{TargetOnly}
        \label{targetOnly_tsne}
    \end{subfigure}
    \begin{subfigure}[b]{0.32\linewidth}
        \centering
        \includegraphics[width=\linewidth]{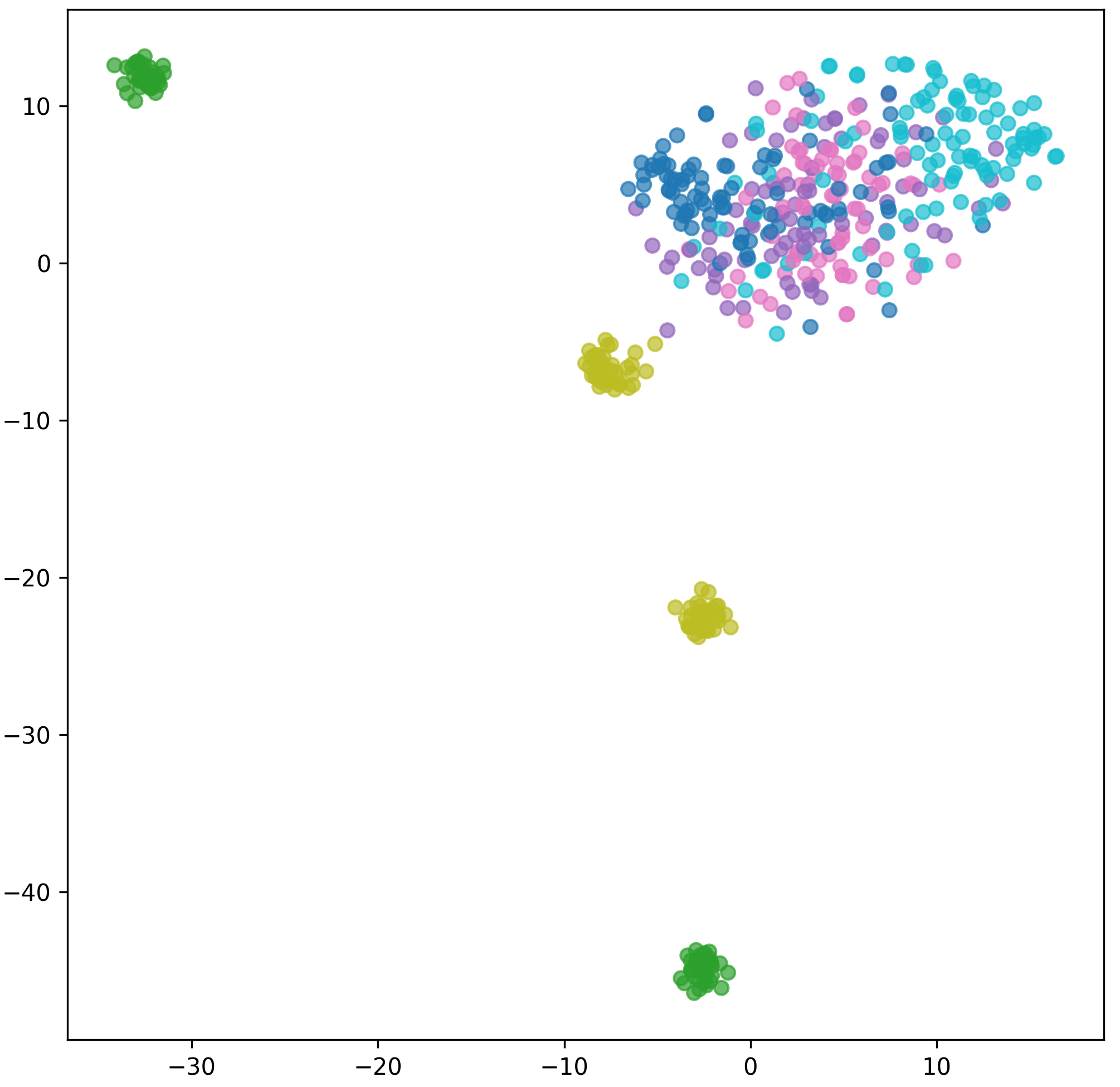}
        \caption{AdaMatch}
        \label{AdaMatch_tSNE}
    \end{subfigure}
    \begin{subfigure}[b]{0.32\linewidth}
        \centering
        \includegraphics[width=\linewidth]{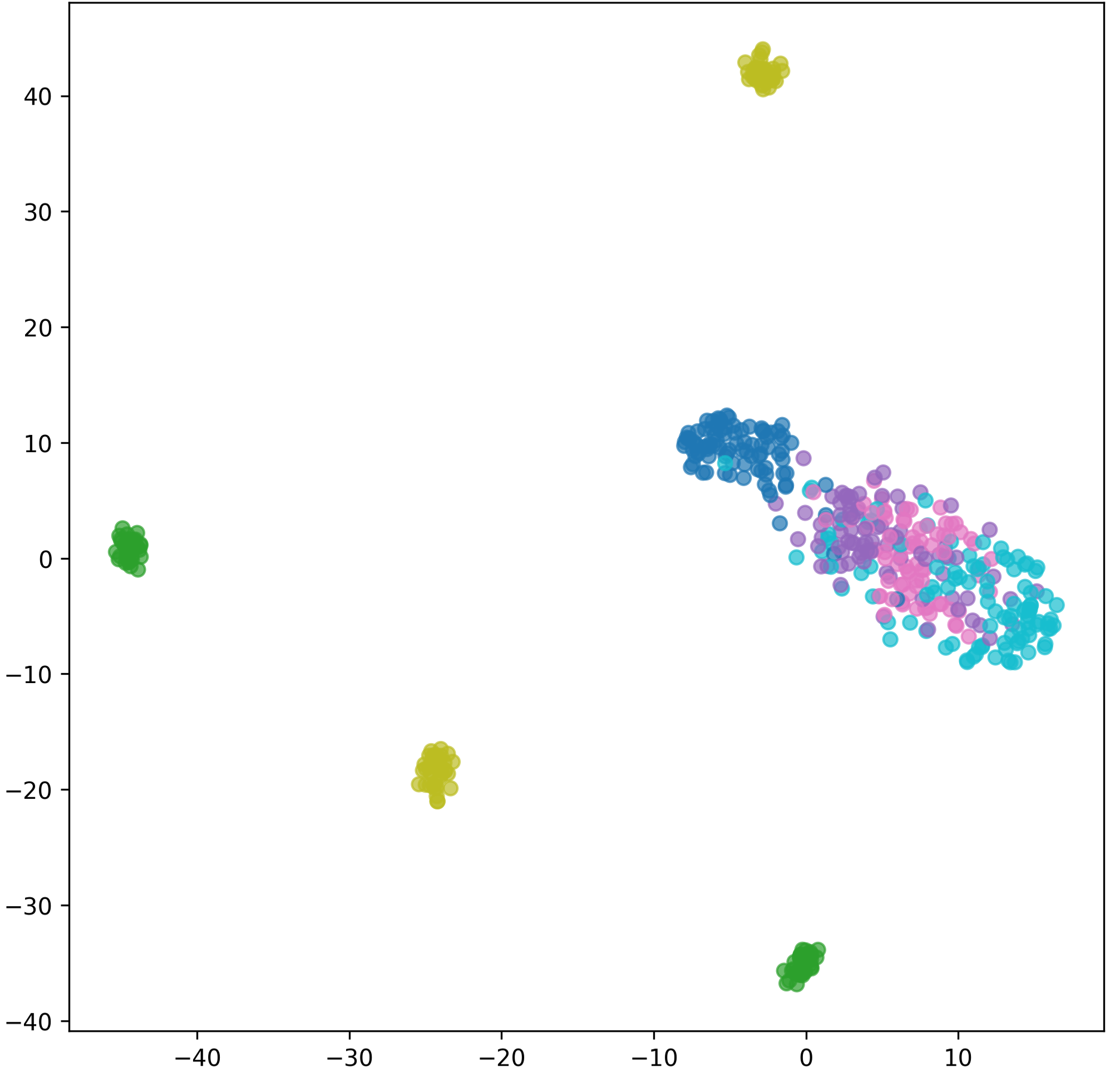}
        \caption{CLDA}
        \label{CLDA_tSNE}
    \end{subfigure}
    \\
    \begin{subfigure}[b]{0.32\linewidth}
        \centering
        \includegraphics[width=\linewidth]{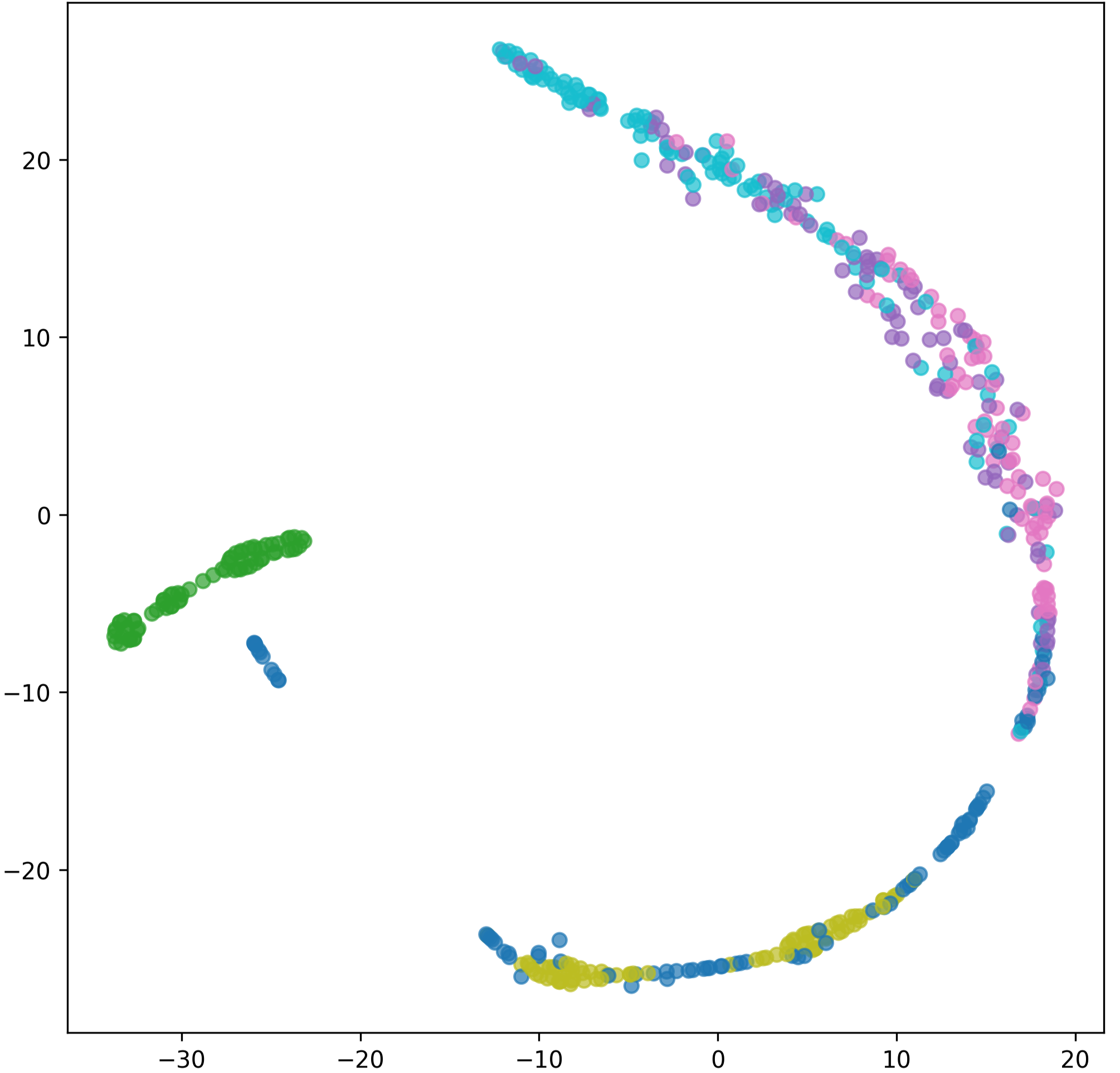}
        \caption{PAC}
        \label{PAC_tSNE}
    \end{subfigure}
    \begin{subfigure}[b]{0.32\linewidth}
        \centering
        \includegraphics[width=\linewidth]{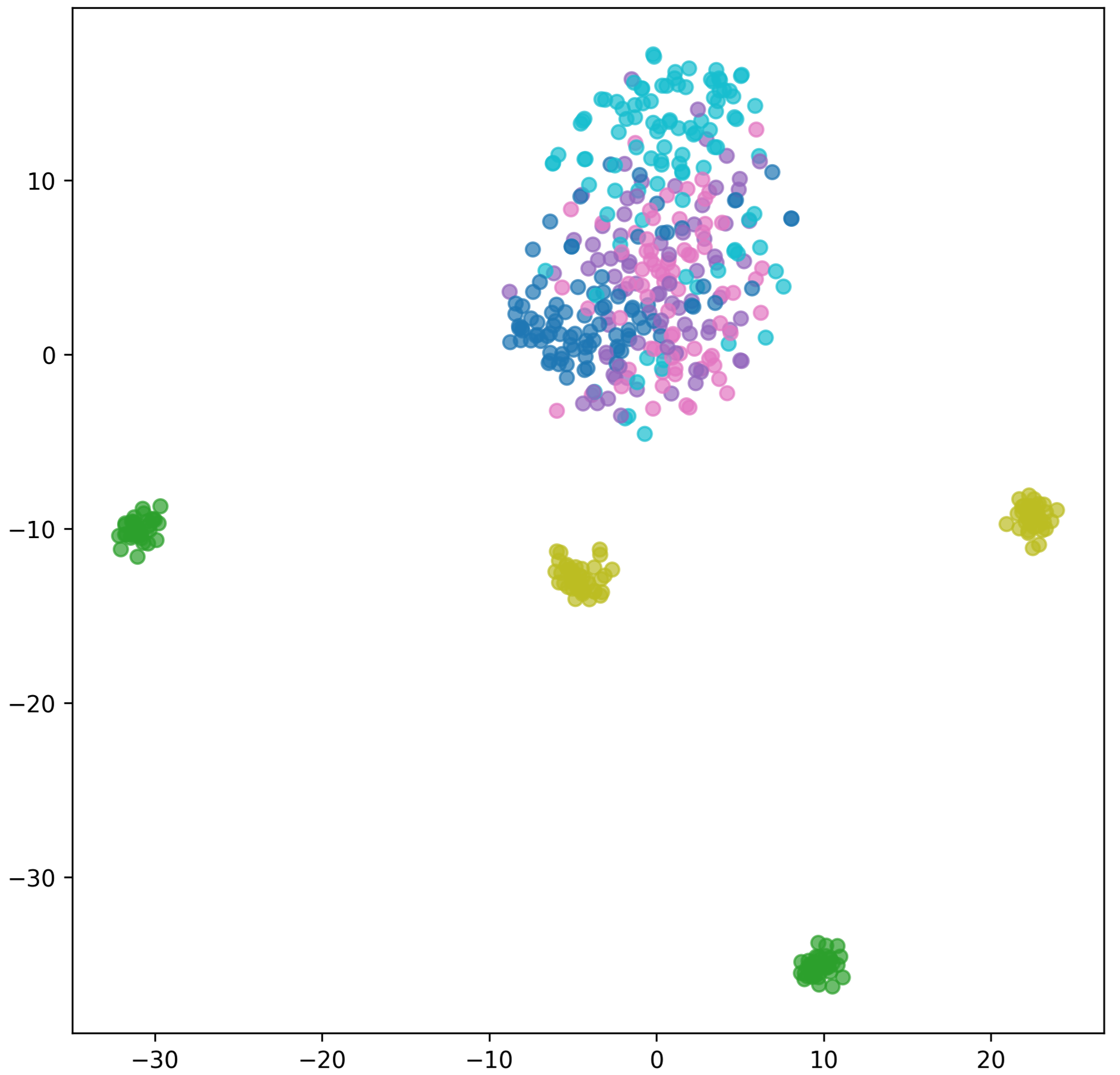}
        \caption{UniSSDA}
        \label{UniSSDA_tSNE}
    \end{subfigure}
    \begin{subfigure}[b]{0.32\linewidth}
        \centering
        \includegraphics[width=\linewidth]{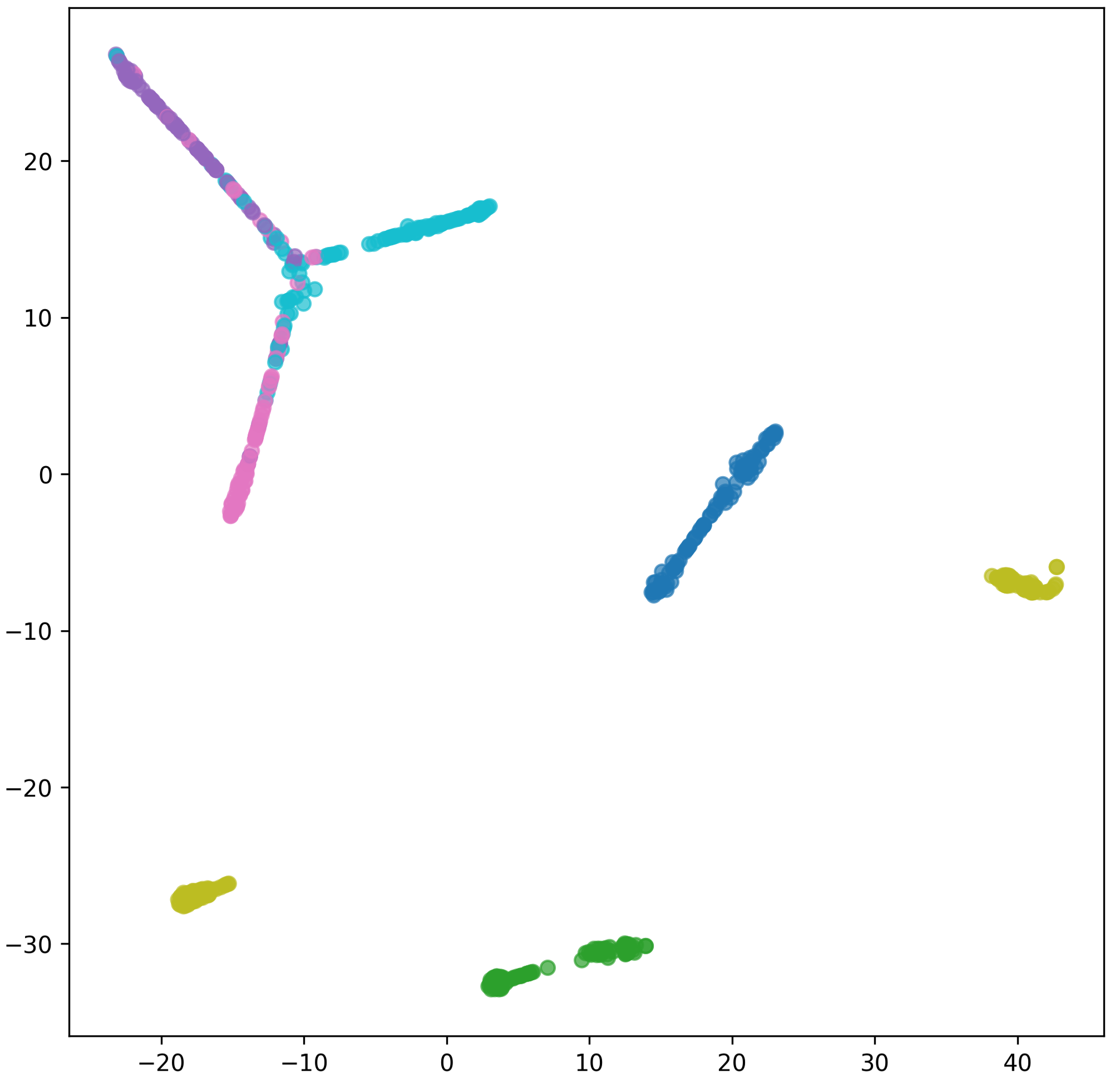}
        \caption{MoSSDA}
        \label{MoSSDA_tSNE}
    \end{subfigure}
\end{minipage}%
\begin{minipage}[c][0.1\linewidth][c]{0.12\linewidth}
    \includegraphics[width=\linewidth]{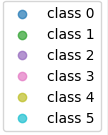}
\end{minipage}
\caption{t-SNE visualization learned on HHAR 1 to 6 DA scenario target test data, when using RESNET18 backbone and unlabeled ratio is 0.95}
\label{tsneHHAR}
\end{figure}

Figure \ref{tsneHHAR} presents a visualization of the feature representations learned using each SSDA method. The visualized features were extracted from the previous layer of each model classifier to ensure consistent comparison. 
For reference, the target-only model—trained with fully labeled target domain data—serves as an upper-bound representation. In cases where only 5\% of the target domain labels were available, our proposed method, MoSSDA method showed the closest alignment to the fully supervised target-only representation. Most benchmark methods failed to separate Class 0 features well, except for CLDA and MoSSDA. MoSSDA achieved clear separation across all six classes, whereas the other methods could not distinguish between classes 2, 3, and 5. These results suggest that MoSSDA effectively enhances the feature discriminability in the target domain under limited label supervision by leveraging labeled and unlabeled target data.

\subsection{Ablation Study} \label{subsec:ablation}

\begin{table}[b] 
\centering

\resizebox{\linewidth}{0.15\linewidth}{%
\begin{tabular}{ccccc|cc|cc} 
\hline\hline
\multicolumn{5}{c}{} & \multicolumn{2}{c}{\rule{0pt}{10pt}\textbf{accuracy}} & \multicolumn{2}{c}{\textbf{f1-score}} \\
~ & \rule{0pt}{11pt}$\mathcal{L}_{mmd}$ & $\mathcal{L}_{ctr}$ & mixup & 2-step & Avg. & diff & Avg. & diff \\ 
\hline
\rule{0pt}{12pt}Proposed & \checkmark & \checkmark & \checkmark & \checkmark & 0.7361 & - & 0.6179 & - \\ \hdashline
\rule{0pt}{11pt}w/o mmd loss &  & \checkmark & \checkmark & \checkmark & 0.7338  & $\downarrow$ 0.002 & 0.6071 & $\downarrow$ 0.011 \\
\rule{0pt}{11pt}w/o ctr loss & \checkmark &  &  & \checkmark & 0.5147 & $\downarrow$ 0.221 & 0.1352 & $\downarrow$ 0.483 \\
\rule{0pt}{11pt}w/o phase1 mix & \checkmark & \checkmark &  & \checkmark & 0.7188 & $\downarrow$ 0.017 & 0.5755  & $\downarrow$ 0.042 \\
\rule{0pt}{11pt}w/o 2-stage learning & \checkmark & \checkmark & \checkmark & ~ & 0.6457 & $\downarrow$ 0.090 & 0.5041  & $\downarrow$ 0.114 \\ 
\hline\hline
\end{tabular}%
}
\caption{Ablation study on effectiveness of proposed methods, evaluated on PTBXL. The unlabeled ratio is 0.7 and backbone is RESNET18}
\label{ablationPTBXL}
\end{table}

\begin{table}[h] 
\centering
\resizebox{\linewidth}{0.15\linewidth}{%
\begin{tabular}{ccccc|cc|cc} 
\hline\hline
\multicolumn{5}{c}{} & \multicolumn{2}{c}{\rule{0pt}{10pt}\textbf{accuracy}} & \multicolumn{2}{c}{\textbf{f1-score}} \\
~ & \rule{0pt}{11pt}$\mathcal{L}_{mmd}$ & $\mathcal{L}_{ctr}$ & mixup & 2-step & Avg. & diff & Avg. & diff \\ 
\hline
\rule{0pt}{12pt}Proposed & \checkmark & \checkmark & \checkmark & \checkmark& 0.9798   & -& 0.9736  & -\\
\hdashline
\rule{0pt}{11pt}w/o mmd loss &  & \checkmark & \checkmark & \checkmark  & 0.9710 & $\downarrow$ 0.009 & 0.9172 & $\downarrow$ 0.056 \\
\rule{0pt}{11pt}w/o ctr loss & \checkmark &  &  & \checkmark & 0.4550 & $\downarrow$ 0.525 & 0.2085 & $\downarrow$ 0.765 \\
\rule{0pt}{11pt}w/o phase1 mix & \checkmark & \checkmark &  & \checkmark & 0.8635 & $\downarrow$ 0.116 & 0.8472 & $\downarrow$ 0.126 \\
\rule{0pt}{11pt}w/o 2-stage learning & \checkmark & \checkmark & \checkmark & ~ & 0.8063 & $\downarrow$ 0.174 & 0.7805 & $\downarrow$ 0.193 \\
\hline\hline
\end{tabular}%
}
\caption{Ablation study on effectiveness of proposed methods, evaluated on MFD. The unlabeled ratio is 0.95 and backbone is CNN}
\label{ablationMFD}
\end{table}

Tables \ref{ablationPTBXL} and \ref{ablationMFD} present the ablation studies on two representative datasets: PTBXL (multivariate, using RESNET18 with an unlabeled ratio of 0.7) and MFD (univariate, using CNN with an unlabeled ratio of 0.95).
The term "diff" denotes the performance gap between MoSSDA and its ablated variants. The exclusion of positive contrastive loss resulted in the most significant performance degradation, with a substantial decline observed, particularly in the F1-score. The 2-stage learning ablation demonstrates the second-largest performance reduction. This result confirms the efficacy of the decoupled learning: the first stage to find class-discriminative representations across intra- and inter-domains, and the second stage to train the classifier with the learned feature space. Mixup ablation revealed that domain-cross mixup facilitates positive contrastive learning by leveraging limited labeled target domain data in domain adaptation scenarios. Similarly, ablating the MMD loss results in performance degradation owing to its failure to minimize the distance between the source and target domains within the backbone architecture.

Across both experiments, the positive contrastive module contributed the most to the overall adaptation performance, followed by the two-step decoupled learning framework. In both experiments, the positive contrastive module contributed the most to the overall adaptation performance, followed by the two-step decoupled learning framework. The impact of additional components, such as mixup within the contrastive module and the domain-invariant encoder is measurable. The F1-score showed larger performance drops than the accuracy when the components were removed, highlighting the robustness of our model in handling class imbalance.

\section{Conclusion} \label{conclusion}
In this study, we propose MoSSDA, a novel framework for the domain adaptation problem within the context of semi-supervised learning for multivariate time series classification.
The MoSSDA features a simple yet effective decoupled learnable structure.
Our approach combines mixup and positive contrastive learning, allowing the model to distinguish between discriminative and consistent features despite highly limited annotations. This architecture incorporates a momentum encoder to ensure the stability and consistency of the learned feature representations, which is a critical factor for time-series data.
The decoupled two-stage learning strategy improves the model robustness and generalization. In addition, our framework allows flexible integration of various backbone models.
The proposed method outperformed state-of-the-art benchmark methods, including augmentation-based SSDA approaches, in highly unlabeled target domain scenarios. Extensive experiments on benchmark datasets validate its superiority. 


\section*{Acknowledgments}
This work was supported by the National Research Foundation of Korea (NRF) grant funded by the Korea government (MSIT) (RS-2024-00455158). This work was also supported by the Technology Innovation Program (20023466, Development of Process Optimization Technology based on Life Cycle Data in Machining Process) funded By the Ministry of Trade, Industry $\&$ Energy (MOTIE, Korea) and Institute of Information $\&$ communications Technology Planning $\&$ Evaluation (IITP) grant funded by the Korea government(MSIT) (No. RS-2022-00155966, Artificial Intelligence Convergence Innovation Human Resources Development (Ewha Womans University)).

\bigskip

\begin{thebibliography}{51}
\providecommand{\natexlab}[1]{#1}

\bibitem[{Anguita et~al.(2013)Anguita, Ghio, Oneto, Parra, Reyes-Ortiz et~al.}]{anguita2013UCIHAR}
Anguita, D.; Ghio, A.; Oneto, L.; Parra, X.; Reyes-Ortiz, J.~L.; et~al. 2013.
\newblock A public domain dataset for human activity recognition using smartphones.
\newblock In \emph{Esann}, volume~3, 3--4.

\bibitem[{Bai, Kolter, and Koltun(2018)}]{bai2018empiricalTCN1}
Bai, S.; Kolter, J.~Z.; and Koltun, V. 2018.
\newblock An empirical evaluation of generic convolutional and recurrent networks for sequence modeling.
\newblock \emph{arXiv preprint arXiv:1803.01271}.

\bibitem[{Berthelot et~al.(2021)Berthelot, Roelofs, Sohn, Carlini, and Kurakin}]{berthelot2021adamatch}
Berthelot, D.; Roelofs, R.; Sohn, K.; Carlini, N.; and Kurakin, A. 2021.
\newblock Adamatch: A unified approach to semi-supervised learning and domain adaptation.
\newblock \emph{arXiv preprint arXiv:2106.04732}.

\bibitem[{Chang et~al.(2024)Chang, Chan, Wang, Peng, and Chen}]{chang2024timedrl}
Chang, C.; Chan, C.-T.; Wang, W.-Y.; Peng, W.-C.; and Chen, T.-F. 2024.
\newblock TimeDRL: Disentangled Representation Learning for Multivariate Time-Series.
\newblock In \emph{2024 IEEE 40th International Conference on Data Engineering (ICDE)}, 625--638. IEEE.

\bibitem[{Chang et~al.(2020)Chang, Mathur, Isopoussu, Song, and Kawsar}]{chang2020systematic1}
Chang, Y.; Mathur, A.; Isopoussu, A.; Song, J.; and Kawsar, F. 2020.
\newblock A systematic study of unsupervised domain adaptation for robust human-activity recognition.
\newblock \emph{Proceedings of the ACM on Interactive, Mobile, Wearable and Ubiquitous Technologies}, 4(1): 1--30.

\bibitem[{Chen et~al.(2022)Chen, Jiang, Wang, Wan, Wang, and Long}]{chen2022DST}
Chen, B.; Jiang, J.; Wang, X.; Wan, P.; Wang, J.; and Long, M. 2022.
\newblock Debiased self-training for semi-supervised learning.
\newblock \emph{Advances in Neural Information Processing Systems}, 35: 32424--32437.

\bibitem[{Chen and He(2021)}]{chen2021noNegative}
Chen, X.; and He, K. 2021.
\newblock Exploring simple siamese representation learning.
\newblock In \emph{Proceedings of the IEEE/CVF conference on computer vision and pattern recognition}, 15750--15758.

\bibitem[{Chen et~al.(2024)Chen, Yan, Yang, Zhang, Zhang, Pan, and Li}]{chen2024CADT}
Chen, Y.; Yan, X.; Yang, Y.; Zhang, J.; Zhang, J.; Pan, L.; and Li, J. 2024.
\newblock Disentangling domain and general representations for time series classification.
\newblock In \emph{Proceedings of the Thirty-Third International Joint Conference on Artificial Intelligence}, 3834--3842.

\bibitem[{Cheng and Pan(2014)}]{cheng2014SSDAmanifold}
Cheng, L.; and Pan, S.~J. 2014.
\newblock Semi-supervised domain adaptation on manifolds.
\newblock \emph{IEEE transactions on neural networks and learning systems}, 25(12): 2240--2249.

\bibitem[{Darban et~al.(2024)Darban, Yang, Webb, Aggarwal, Wen, and Salehi}]{darban2024dacad}
Darban, Z.~Z.; Yang, Y.; Webb, G.~I.; Aggarwal, C.~C.; Wen, Q.; and Salehi, M. 2024.
\newblock DACAD: Domain adaptation contrastive learning for anomaly detection in multivariate time series.
\newblock \emph{arXiv preprint arXiv:2404.11269}.

\bibitem[{Deng, Tu, and Xu(2021)}]{deng2021multiUDAecg}
Deng, F.; Tu, S.; and Xu, L. 2021.
\newblock Multi-source unsupervised domain adaptation for ECG classification.
\newblock In \emph{2021 IEEE International Conference on Bioinformatics and Biomedicine (BIBM)}, 854--859. IEEE.

\bibitem[{Eldele et~al.(2023)Eldele, Ragab, Chen, Wu, Kwoh, and Li}]{eldele2023CoTMix}
Eldele, E.; Ragab, M.; Chen, Z.; Wu, M.; Kwoh, C.-K.; and Li, X. 2023.
\newblock Contrastive domain adaptation for time-series via temporal mixup.
\newblock \emph{IEEE Transactions on Artificial Intelligence}, 5(3): 1185--1194.

\bibitem[{Eldele et~al.(2021{\natexlab{a}})Eldele, Ragab, Chen, Wu, Kwoh, Li, and Guan}]{eldele2021time2}
Eldele, E.; Ragab, M.; Chen, Z.; Wu, M.; Kwoh, C.~K.; Li, X.; and Guan, C. 2021{\natexlab{a}}.
\newblock Time-series representation learning via temporal and contextual contrasting.
\newblock \emph{arXiv preprint arXiv:2106.14112}.

\bibitem[{Eldele et~al.(2021{\natexlab{b}})Eldele, Ragab, Chen, Wu, Kwoh, Li, and Guan}]{eldele2021timeCNN1}
Eldele, E.; Ragab, M.; Chen, Z.; Wu, M.; Kwoh, C.~K.; Li, X.; and Guan, C. 2021{\natexlab{b}}.
\newblock Time-series representation learning via temporal and contextual contrasting.
\newblock \emph{arXiv preprint arXiv:2106.14112}.

\bibitem[{Eldele et~al.(2022)Eldele, Ragab, Chen, Wu, Kwoh, Li, and Guan}]{eldele2022adastCNN2}
Eldele, E.; Ragab, M.; Chen, Z.; Wu, M.; Kwoh, C.-K.; Li, X.; and Guan, C. 2022.
\newblock ADAST: Attentive cross-domain EEG-based sleep staging framework with iterative self-training.
\newblock \emph{IEEE Transactions on Emerging Topics in Computational Intelligence}, 7(1): 210--221.

\bibitem[{Fawaz(2020)}]{fawaz2020deepRESNET18}
Fawaz, H.~I. 2020.
\newblock Deep learning for time series classification.
\newblock \emph{arXiv preprint arXiv:2010.00567}.

\bibitem[{Ganin and Lempitsky(2015)}]{ganin2015adversarialUDA}
Ganin, Y.; and Lempitsky, V. 2015.
\newblock Unsupervised domain adaptation by backpropagation.
\newblock In \emph{International conference on machine learning}, 1180--1189. PMLR.

\bibitem[{Goldberger et~al.(2000)Goldberger, Amaral, Glass, Hausdorff, Ivanov, Mark, Mietus, Moody, Peng, and Stanley}]{goldberger2000EEG}
Goldberger, A.~L.; Amaral, L.~A.; Glass, L.; Hausdorff, J.~M.; Ivanov, P.~C.; Mark, R.~G.; Mietus, J.~E.; Moody, G.~B.; Peng, C.-K.; and Stanley, H.~E. 2000.
\newblock PhysioBank, PhysioToolkit, and PhysioNet: components of a new research resource for complex physiologic signals.
\newblock \emph{circulation}, 101(23): e215--e220.

\bibitem[{Grill et~al.(2020)Grill, Strub, Altch{\'e}, Tallec, Richemond, Buchatskaya, Doersch, Avila~Pires, Guo, Gheshlaghi~Azar et~al.}]{grill2020BYOL}
Grill, J.-B.; Strub, F.; Altch{\'e}, F.; Tallec, C.; Richemond, P.; Buchatskaya, E.; Doersch, C.; Avila~Pires, B.; Guo, Z.; Gheshlaghi~Azar, M.; et~al. 2020.
\newblock Bootstrap your own latent-a new approach to self-supervised learning.
\newblock \emph{Advances in neural information processing systems}, 33: 21271--21284.

\bibitem[{He et~al.(2023)He, Queen, Koker, Cuevas, Tsiligkaridis, and Zitnik}]{he2023domain}
He, H.; Queen, O.; Koker, T.; Cuevas, C.; Tsiligkaridis, T.; and Zitnik, M. 2023.
\newblock Domain adaptation for time series under feature and label shifts.
\newblock In \emph{International conference on machine learning}, 12746--12774. PMLR.

\bibitem[{He et~al.(2020)He, Fan, Wu, Xie, and Girshick}]{he2020MoCo}
He, K.; Fan, H.; Wu, Y.; Xie, S.; and Girshick, R. 2020.
\newblock Momentum contrast for unsupervised visual representation learning.
\newblock In \emph{Proceedings of the IEEE/CVF conference on computer vision and pattern recognition}, 9729--9738.

\bibitem[{He et~al.(2016)He, Zhang, Ren, and Sun}]{he2016deepRESNET18}
He, K.; Zhang, X.; Ren, S.; and Sun, J. 2016.
\newblock Deep residual learning for image recognition.
\newblock In \emph{Proceedings of the IEEE conference on computer vision and pattern recognition}, 770--778.

\bibitem[{Iglesias et~al.(2023)Iglesias, Talavera, Gonz{\'a}lez-Prieto, Mozo, and G{\'o}mez-Canaval}]{iglesias2023augTS1}
Iglesias, G.; Talavera, E.; Gonz{\'a}lez-Prieto, {\'A}.; Mozo, A.; and G{\'o}mez-Canaval, S. 2023.
\newblock Data augmentation techniques in time series domain: a survey and taxonomy.
\newblock \emph{Neural Computing and Applications}, 35(14): 10123--10145.

\bibitem[{Ilbert, Hoang, and Zhang(2024)}]{ilbert2024augTS2}
Ilbert, R.; Hoang, T.~V.; and Zhang, Z. 2024.
\newblock Data augmentation for multivariate time series classification: An experimental study.
\newblock In \emph{2024 IEEE 40th International Conference on Data Engineering Workshops (ICDEW)}, 128--139. IEEE.

\bibitem[{Jin et~al.(2022)Jin, Park, Maddix, Wang, and Wang}]{jin2022domain}
Jin, X.; Park, Y.; Maddix, D.; Wang, H.; and Wang, Y. 2022.
\newblock Domain adaptation for time series forecasting via attention sharing.
\newblock In \emph{International Conference on Machine Learning}, 10280--10297. PMLR.

\bibitem[{Khosla et~al.(2020)Khosla, Teterwak, Wang, Sarna, Tian, Isola, Maschinot, Liu, and Krishnan}]{khosla2020supervisedCL}
Khosla, P.; Teterwak, P.; Wang, C.; Sarna, A.; Tian, Y.; Isola, P.; Maschinot, A.; Liu, C.; and Krishnan, D. 2020.
\newblock Supervised contrastive learning.
\newblock \emph{Advances in neural information processing systems}, 33: 18661--18673.

\bibitem[{Kim et~al.(2022)Kim, Ngo, Park, Kwon, Lee, and Cho}]{kim2022DARK}
Kim, J.~H.; Ngo, B.~H.; Park, J.~H.; Kwon, J.~E.; Lee, H.~S.; and Cho, S.~I. 2022.
\newblock Distilling and Refining Domain-Specific Knowledge for Semi-Supervised Domain Adaptation.
\newblock In \emph{BMVC}, 606.

\bibitem[{Kwapisz, Weiss, and Moore(2011)}]{kwapisz2011activityWISDM}
Kwapisz, J.~R.; Weiss, G.~M.; and Moore, S.~A. 2011.
\newblock Activity recognition using cell phone accelerometers.
\newblock \emph{ACM SigKDD Explorations Newsletter}, 12(2): 74--82.

\bibitem[{Lessmeier et~al.(2016)Lessmeier, Kimotho, Zimmer, and Sextro}]{lessmeier2016MFD}
Lessmeier, C.; Kimotho, J.~K.; Zimmer, D.; and Sextro, W. 2016.
\newblock Condition monitoring of bearing damage in electromechanical drive systems by using motor current signals of electric motors: A benchmark data set for data-driven classification.
\newblock In \emph{PHM society European conference}, volume~3.

\bibitem[{Li et~al.(2021{\natexlab{a}})Li, Li, Shi, and Yu}]{li2021CDAC}
Li, J.; Li, G.; Shi, Y.; and Yu, Y. 2021{\natexlab{a}}.
\newblock Cross-domain adaptive clustering for semi-supervised domain adaptation.
\newblock In \emph{Proceedings of the IEEE/CVF conference on computer vision and pattern recognition}, 2505--2514.

\bibitem[{Li et~al.(2021{\natexlab{b}})Li, Liu, Zhao, Zhang, and Fu}]{li2021ecacl}
Li, K.; Liu, C.; Zhao, H.; Zhang, Y.; and Fu, Y. 2021{\natexlab{b}}.
\newblock Ecacl: A holistic framework for semi-supervised domain adaptation.
\newblock In \emph{Proceedings of the IEEE/CVF international conference on computer vision}, 8578--8587.

\bibitem[{Li et~al.(2021{\natexlab{c}})Li, Li, Chen, Zhou, Zeng, and Li}]{li2021modeling3}
Li, Y.; Li, K.; Chen, C.; Zhou, X.; Zeng, Z.; and Li, K. 2021{\natexlab{c}}.
\newblock Modeling temporal patterns with dilated convolutions for time-series forecasting.
\newblock \emph{ACM Transactions on Knowledge Discovery from Data (TKDD)}, 16(1): 1--22.

\bibitem[{Liu and Xue(2021)}]{liu2021AdvSKM}
Liu, Q.; and Xue, H. 2021.
\newblock Adversarial Spectral Kernel Matching for Unsupervised Time Series Domain Adaptation.
\newblock In \emph{IJCAI}, 2744--2750.

\bibitem[{Long et~al.(2018)Long, Cao, Wang, and Jordan}]{long2018CDAN}
Long, M.; Cao, Z.; Wang, J.; and Jordan, M.~I. 2018.
\newblock Conditional adversarial domain adaptation.
\newblock \emph{Advances in neural information processing systems}, 31.

\bibitem[{Mishra, Saenko, and Saligrama(2021)}]{mishra2021PAC}
Mishra, S.; Saenko, K.; and Saligrama, V. 2021.
\newblock Surprisingly simple semi-supervised domain adaptation with pretraining and consistency.
\newblock \emph{arXiv preprint arXiv:2101.12727}.

\bibitem[{Ott et~al.(2022)Ott, R{\"u}gamer, Heublein, Bischl, and Mutschler}]{ott2022domainLabelShift}
Ott, F.; R{\"u}gamer, D.; Heublein, L.; Bischl, B.; and Mutschler, C. 2022.
\newblock Domain adaptation for time-series classification to mitigate covariate shift.
\newblock In \emph{Proceedings of the 30th ACM international conference on multimedia}, 5934--5943.

\bibitem[{Ragab et~al.(2023)Ragab, Eldele, Tan, Foo, Chen, Wu, Kwoh, and Li}]{ragab2023adatime}
Ragab, M.; Eldele, E.; Tan, W.~L.; Foo, C.-S.; Chen, Z.; Wu, M.; Kwoh, C.-K.; and Li, X. 2023.
\newblock Adatime: A benchmarking suite for domain adaptation on time series data.
\newblock \emph{ACM Transactions on Knowledge Discovery from Data}, 17(8): 1--18.

\bibitem[{Saito et~al.(2019)Saito, Kim, Sclaroff, Darrell, and Saenko}]{saito2019minimaxSSDA}
Saito, K.; Kim, D.; Sclaroff, S.; Darrell, T.; and Saenko, K. 2019.
\newblock Semi-supervised domain adaptation via minimax entropy.
\newblock In \emph{Proceedings of the IEEE/CVF international conference on computer vision}, 8050--8058.

\bibitem[{Shi, Ying, and Yang(2022)}]{shi2022deepSurvey}
Shi, Y.; Ying, X.; and Yang, J. 2022.
\newblock Deep unsupervised domain adaptation with time series sensor data: A survey.
\newblock \emph{Sensors}, 22(15): 5507.

\bibitem[{Shu et~al.(2018)Shu, Bui, Narui, and Ermon}]{shu2018dirt}
Shu, R.; Bui, H.~H.; Narui, H.; and Ermon, S. 2018.
\newblock A dirt-t approach to unsupervised domain adaptation.
\newblock \emph{arXiv preprint arXiv:1802.08735}.

\bibitem[{Singh(2021)}]{singh2021clda}
Singh, A. 2021.
\newblock Clda: Contrastive learning for semi-supervised domain adaptation.
\newblock \emph{Advances in neural information processing systems}, 34: 5089--5101.

\bibitem[{Stisen et~al.(2015)Stisen, Blunck, Bhattacharya, Prentow, Kj{\ae}rgaard, Dey, Sonne, and Jensen}]{stisen2015HHAR}
Stisen, A.; Blunck, H.; Bhattacharya, S.; Prentow, T.~S.; Kj{\ae}rgaard, M.~B.; Dey, A.; Sonne, T.; and Jensen, M.~M. 2015.
\newblock Smart devices are different: Assessing and mitigatingmobile sensing heterogeneities for activity recognition.
\newblock In \emph{Proceedings of the 13th ACM conference on embedded networked sensor systems}, 127--140.

\bibitem[{Sun et~al.(2024)}]{sun2024caudits}
Sun, S.; et~al. 2024.
\newblock CaudiTS: Causal disentangled domain adaptation of multivariate time series.
\newblock In \emph{Forty-first International Conference on Machine Learning}.

\bibitem[{Thill, Konen, and B{\"a}ck(2020)}]{thill2020timeTCN2}
Thill, M.; Konen, W.; and B{\"a}ck, T. 2020.
\newblock Time series encodings with temporal convolutional networks.
\newblock In \emph{International Conference on Bioinspired Methods and Their Applications}, 161--173. Springer.

\bibitem[{Wagner et~al.(2022)Wagner, Strodthoff, Bousseljot, Samek, and Schaeffter}]{wagner10ptbxl}
Wagner, P.; Strodthoff, N.; Bousseljot, R.; Samek, W.; and Schaeffter, T. 2022.
\newblock PTB-XL, a large publicly available electrocardiography dataset (version 1.0. 3), 2022.
\newblock \emph{URL https://doi. org/10.13026/kfzx-aw45}.

\bibitem[{Wagner et~al.(2020)Wagner, Strodthoff, Bousseljot, Kreiseler, Lunze, Samek, and Schaeffter}]{wagner2020ptbxl}
Wagner, P.; Strodthoff, N.; Bousseljot, R.-D.; Kreiseler, D.; Lunze, F.~I.; Samek, W.; and Schaeffter, T. 2020.
\newblock PTB-XL, a large publicly available electrocardiography dataset.
\newblock \emph{Scientific data}, 7(1): 1--15.

\bibitem[{Wilson, Doppa, and Cook(2020)}]{wilson2020CoDATS}
Wilson, G.; Doppa, J.~R.; and Cook, D.~J. 2020.
\newblock Multi-source deep domain adaptation with weak supervision for time-series sensor data.
\newblock In \emph{Proceedings of the 26th ACM SIGKDD international conference on knowledge discovery \& data mining}, 1768--1778.

\bibitem[{Yang et~al.(2021)Yang, Wang, Gao, Shrivastava, Weinberger, Chao, and Lim}]{yang2021deepCoTraining}
Yang, L.; Wang, Y.; Gao, M.; Shrivastava, A.; Weinberger, K.~Q.; Chao, W.-L.; and Lim, S.-N. 2021.
\newblock Deep co-training with task decomposition for semi-supervised domain adaptation.
\newblock In \emph{Proceedings of the IEEE/CVF international conference on computer vision}, 8906--8916.

\bibitem[{Yoon, Kang, and Cho(2022)}]{yoon2022sampletosample}
Yoon, J.; Kang, D.; and Cho, M. 2022.
\newblock Semi-supervised domain adaptation via sample-to-sample self-distillation.
\newblock In \emph{Proceedings of the IEEE/CVF Winter Conference on Applications of Computer Vision}, 1978--1987.

\bibitem[{Zhang et~al.(2017)Zhang, Cisse, Dauphin, and Lopez-Paz}]{zhang2017mixup}
Zhang, H.; Cisse, M.; Dauphin, Y.~N.; and Lopez-Paz, D. 2017.
\newblock mixup: Beyond empirical risk minimization.
\newblock \emph{arXiv preprint arXiv:1710.09412}.

\bibitem[{Zhang et~al.(2024)Zhang, Liu, Cong, Ragab, and Foo}]{zhang2024uniSSDA}
Zhang, W.; Liu, Q.; Cong, F. O.~W.; Ragab, M.; and Foo, C.-S. 2024.
\newblock Universal Semi-Supervised Domain Adaptation by Mitigating Common-Class Bias.
\newblock In \emph{Proceedings of the IEEE/CVF Conference on Computer Vision and Pattern Recognition}, 23912--23921.

\end{thebibliography}

\clearpage
\newpage
\appendix \label{appendix}

\newcommand{\beginsupplement}{%
        \setcounter{table}{0}
        \renewcommand{\thetable}{S\arabic{table}}%
        \setcounter{figure}{0}
        \renewcommand{\thefigure}{S\arabic{figure}}%

     }
\newcounter{suppsubsection}
\newcounter{suppsubsubsection}[suppsubsection]

\newcommand{\suppsubsection}[1]{%
  \refstepcounter{suppsubsection}%
  \paragraph{S.\arabic{suppsubsection}. #1}%
  \setcounter{suppsubsubsection}{0}%
}
\newcommand{\suppsubsubsection}{%
  \refstepcounter{suppsubsubsection}%
  \paragraph{S.\arabic{suppsubsection}.\arabic{suppsubsubsection}}%
}

\section{Supplementary} \label{sec:supp}
\beginsupplement
\suppsubsection{Dataset Details} \label{subsec:dataset}
\suppsubsubsection{UCIHAR} \label{subsubsec:UCIHAR}
\cite{anguita2013UCIHAR} comprises sensor data collected from 30 subjects during six activities: walking, ascending stairs, descending stairs, standing, sitting, and lying down. The data acquisition process incorporated accelerometers, gyroscopes, and body sensors. Each subject is treated as a distinct domain to account for inter-subject variability.

\suppsubsubsection{WISDM}
\cite{kwapisz2011activityWISDM} comprises accelerometer readings from 36 participants who performed the same six physical activities described for \ref{subsubsec:UCIHAR}. This dataset is notable for its ability to capture temporal variability, thereby facilitating the evaluation of domain adaptation methods in activity recognition tasks.

\suppsubsubsection{HHAR}
\cite{stisen2015HHAR} encompasses the sensor signals from nine individuals' smartphones and smartwatches. The data for each subject constitutes a distinct domain, facilitating the exploration of heterogeneity in sensor-based human activity recognition.

\suppsubsubsection{PTBXL}
\cite{wagner10ptbxl,wagner2020ptbxl} is a substantial clinical dataset comprising 12-lead electrocardiogram (ECG) signals. The ECG recordings were obtained from 11 distinct ECG device models, of which the three most represented devices define unique domains. The classification task is organized into five diagnostic super-classes, with significant class imbalance and domain-distribution heterogeneity presenting notable challenges.

\suppsubsubsection{EEG}
 \cite{goldberger2000EEG} under consideration includes single-channel recordings from 20 healthy subjects classified into five sleep stages: Wake, N1, N2, N3, and REM. Each individual's data constitutes a domain, thereby facilitating subject transfer scenarios in sleep stage classification.

\suppsubsubsection{MFD}
 \cite{lessmeier2016MFD} comprises uni-variate vibration signals obtained under four distinct operating conditions, each treated as a distinct domain. The dataset is utilized to evaluate the efficacy of the model in the context of initial fault detection.

\suppsubsection{Benchmark Methods Details}
\suppsubsubsection{AdaMatch}
\cite{berthelot2021adamatch} provides a unified approach for semi-supervised domain adaptation by applying both weak and strong augmentations to achieve effective distribution alignment between source and target data.
\suppsubsubsection{CDAC}
\cite{li2021CDAC} method addresses inter- and intra-domain adaptation by employing adversarial adaptive clustering loss and aligning feature clusters across domains. Pseudo-labeling is utilized to expand the set of labeled samples during training.
\suppsubsubsection{DST}
\cite{chen2022DST} mitigates self-training bias in semi-supervised settings by decoupling pseudo-label generation and utilization across two classifier heads and adversarially optimizing feature representations to improve pseudo-label quality.
\suppsubsubsection{PAC}
\cite{mishra2021PAC} demonstrates that a robust target classifier can be obtained through self-supervised pretraining (e.g., rotation prediction) and consistency regularization, obviating the need for explicit source-target alignment in semi-supervised domain adaptation.
\suppsubsubsection{UniSSDA}
\cite{zhang2024uniSSDA} addresses common-class bias in universal domain adaptation by introducing a prior-guided pseudo-label refinement strategy, supporting mixed private and common class scenarios for both source and target domains.
\suppsubsubsection{CLDA}
\cite{singh2021clda} is a single-stage contrastive learning framework comprising inter-domain contrastive alignment of class centroids and instance-level similarity maximization, thereby enhancing representation learning under semi-supervised domain adaptation

\suppsubsection{Implementation Details} \label{subsec:implementation}

\suppsubsubsection{Model Backbones}
\begin{itemize}
\item \textbf{Convolutional Neural Network (CNN)}: The employed 1D-CNN architecture comprises three convolutional blocks, each integrating a convolutional layer, batch normalization, ReLU activation, and max pooling. The structure has been designed to extract sequential patterns from time series data.
\item \textbf{RESNET18}: ResNet-18 for 1D data incorporates residual connections to facilitate deep network training by enabling information flow across layers. This design has been demonstrated to effectively mitigate vanishing gradient effects while concurrently enhancing the efficacy of feature learning in the context of time series analysis.
\item \textbf{Temporal Convolutional Network (TCN)}: The TCN employs causal, dilated convolutions to capture long-range temporal dependencies in sequential data, while effectively preventing information leakage across temporal blocks.
\end{itemize}

\suppsubsubsection{Augmentations}

Since the data augmentations used in the benchmark methods are intended for images, we replaced them with augmentations appropriate for our multivariate time series implementation.
We implemented a suite of augmentations tailored for time series data.:
\begin{itemize}
\item \textbf{TSRandomHorizontalFlip}: randomly reverses the temporal sequence 50 percent of the time.
\item \textbf{RandomErasingTS}: zero-masks randomly selected segments to improve robustness to missing data.
\item \textbf{RandAugmentTS}: applies random augmentations sequentially from a predefined pool. The number (n) and strength (m) of transformations are controlled.
\item \textbf{AddNoise}: introduces Gaussian noise to the input sequence.
\item \textbf{Scale}: modifies the signal amplitude.
\item \textbf{TimeWarp}: applies nonlinear temporal warping based on a beta distribution.
\item \textbf{Cutout1D}: masks the input sequence by setting values to zero, similar to RandomErasingTS.
\item \textbf{Permute}: segments and shuffles ordered batches of the input sequence.
\end{itemize}
The augmentations are composed differently per phase. One transformation is used during training (n = 1, m = 9); two transformations are used during the strong augmentation phases (n = 2, m = 10); and no transformations are applied during validation or testing.

\begin{figure}[ht]
\centering
\subfloat[MoSSDA test accuracy score for HAR dataset]
{\scalebox{0.55}{\includegraphics{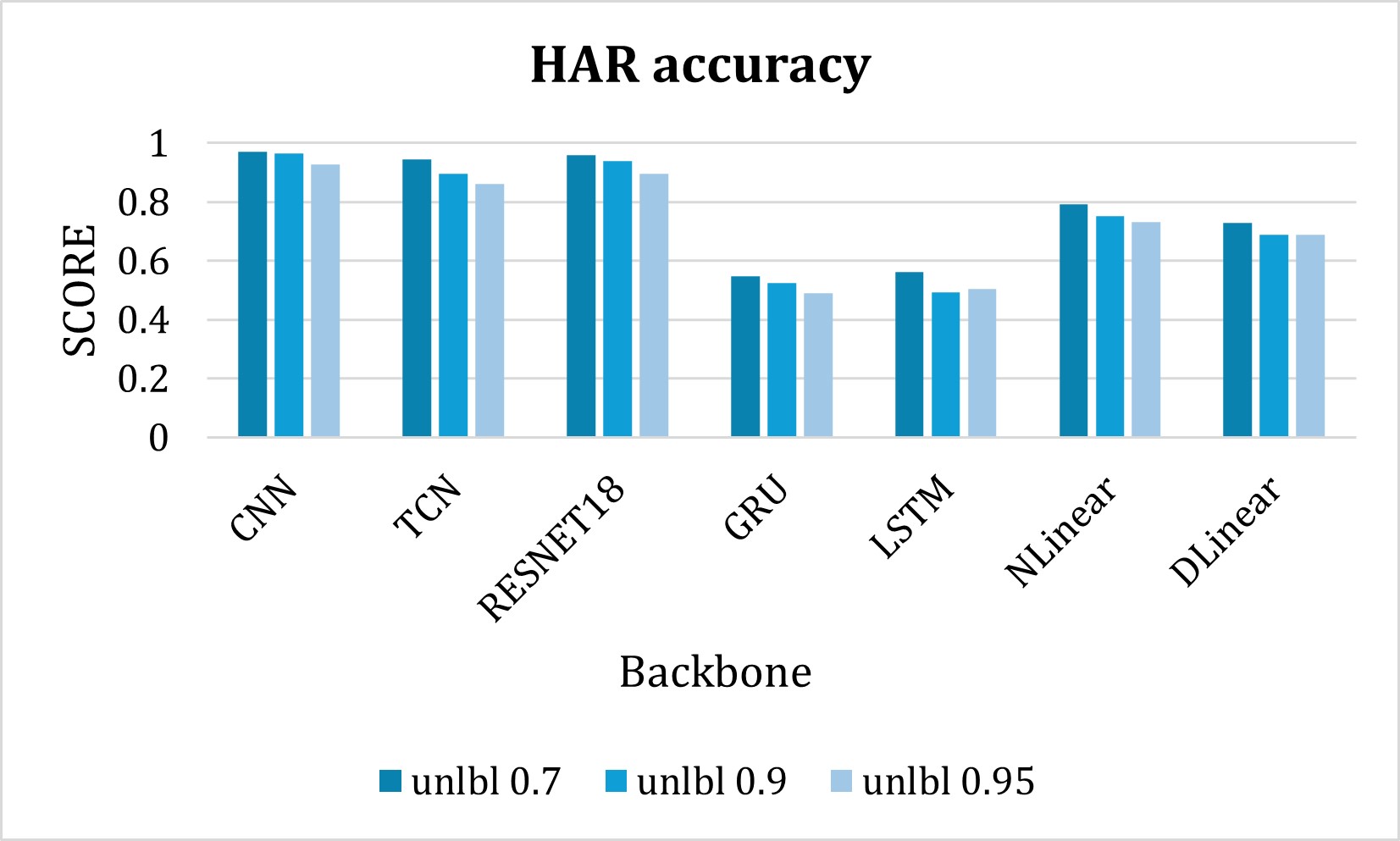}}}
\vfill
\subfloat[MoSSDA test f1-score score for HAR dataset]
{\scalebox{0.55}{\includegraphics{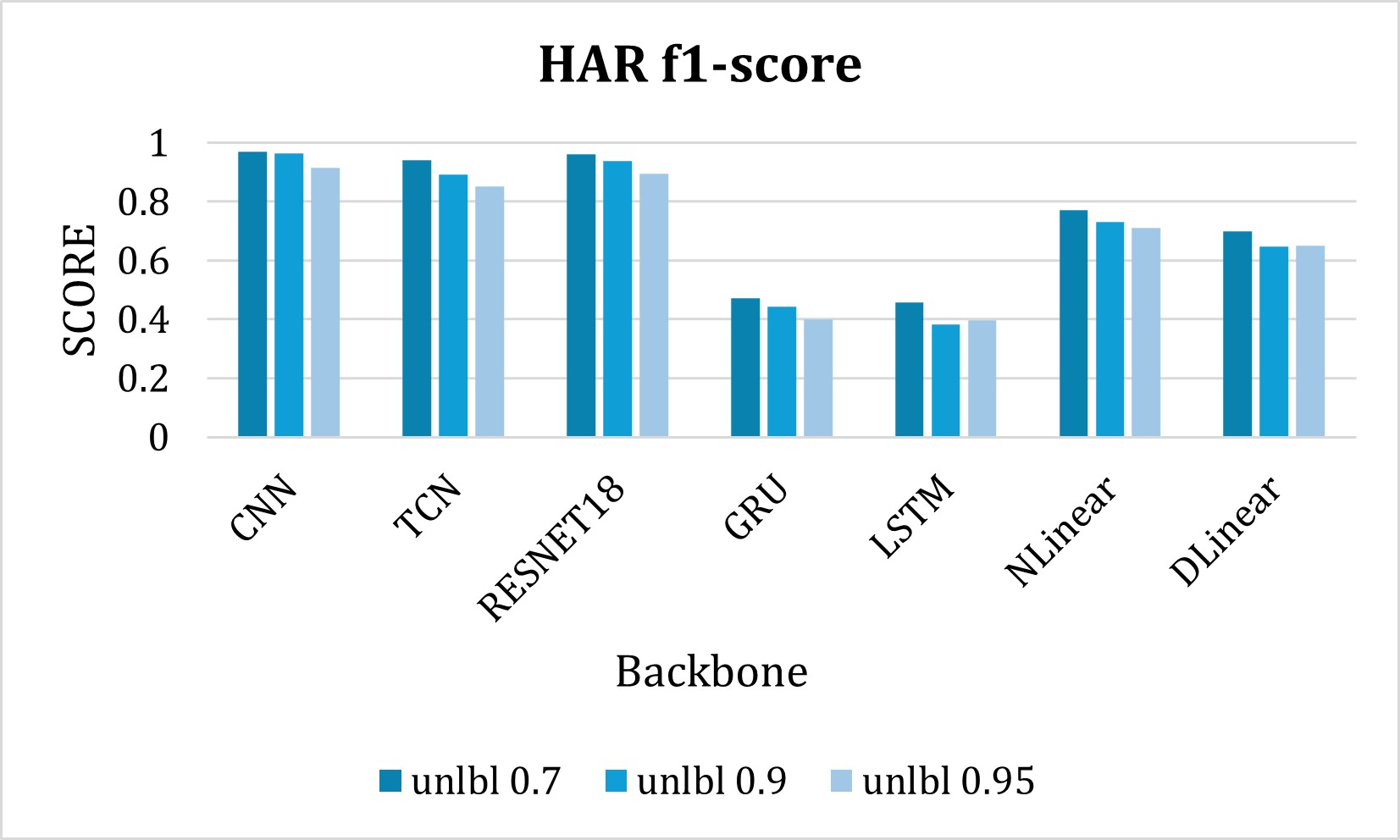}}}
\caption{Comparison with different backbone networks.}
\label{fig:backbones}
\end{figure}

\suppsubsection{Full Results of Backbones}
To further validate our approach, we extended our experiments by incorporating four additional backbones tailored for time-series data. These backbones include two recurrent neural network architectures—GRU and LSTM—and two multi-layer perceptron-based models—NLinear and DLinear. The Figure \ref{fig:backbones} illustrates the performance on the HAR dataset, evaluated on the target domain test set under three different unlabeled data ratios: 0.7, 0.9, and 0.95. This comparison highlights how each backbone performs under increasing scarcity of labeled target data.

\suppsubsection{Full Ablation Study Results}
Tables \ref{ablationHHAR_full}, \ref{ablationPTBXL_full}, \ref{ablationEEG_full}, and \ref{ablationMFD_full} present the results of ablation studies across four different time-series datasets:
PTBXL for electrocardiogram (ECG) based diagnosis
EEG for sleep stage prediction
HHAR for human activity recognition
MFD for machining fault classification
In each case, experiments were performed under three fixed unlabeled ratios (0.7, 0.9, and 0.95), and compared across three backbone architectures—CNN, ResNet18, and TCN. These results provide insight into the contribution of each component in our method and the influence of different backbone choices across various domains.

\begin{table*}[]
\centering
\resizebox{\textwidth}{!}{%
\begin{tabular}{cccccc|cccccccccccc}
\hline
\multicolumn{6}{c|}{} & \multicolumn{4}{c}{\textbf{unlab.\_ratio   = 0.7}} & \multicolumn{4}{c}{\textbf{unlab.\_ratio = 0.9}} & \multicolumn{4}{c}{\textbf{unlab.\_ratio = 0.95}} \\
\multicolumn{6}{c|}{\multirow{-2}{*}{\textbf{}}} & \multicolumn{2}{c}{\textbf{accuracy}} & \multicolumn{2}{c}{\textbf{f1-score}} & \multicolumn{2}{c}{\textbf{accuracy}} & \multicolumn{2}{c}{\textbf{f1-score}} & \multicolumn{2}{c}{\textbf{accuracy}} & \multicolumn{2}{c}{\textbf{f1-score}} \\ \cline{7-18} 
 & mmd\_loss & ctr\_loss & phase1 mix & 2-step & backbone & Avg. & std. & Avg. & std. & Avg. & std. & Avg. & std. & Avg. & std. & Avg. & std. \\ \hline
\rowcolor[gray]{0.9} 
\cellcolor[gray]{0.9} & \cellcolor[gray]{0.9} & \cellcolor[gray]{0.9} & \cellcolor[gray]{0.9} & \cellcolor[gray]{0.9} & CNN & 0.9784 & ±0.01 & 0.9788 & ±0.01 & 0.9595 & ±0.02 & 0.9604 & ±0.02 & 0.9480 & ±0.03 & 0.9494 & ±0.03 \\
\rowcolor[gray]{0.9} 
\cellcolor[gray]{0.9} & \cellcolor[gray]{0.9} & \cellcolor[gray]{0.9} & \cellcolor[gray]{0.9} & \cellcolor[gray]{0.9} & RESNET18 & 0.9693 & ±0.02 & 0.9698 & ±0.02 & 0.9563 & ±0.02 & 0.9567 & ±0.02 & 0.9430 & ±0.03 & 0.9433 & ±0.03 \\
\rowcolor[gray]{0.9} 
\multirow{-3}{*}{\cellcolor[gray]{0.9}Proposed} & \multirow{-3}{*}{\cellcolor[gray]{0.9}\checkmark} & \multirow{-3}{*}{\cellcolor[gray]{0.9}\checkmark} & \multirow{-3}{*}{\cellcolor[gray]{0.9}\checkmark} & \multirow{-3}{*}{\cellcolor[gray]{0.9}\checkmark} & TCN & 0.9089 & ±0.06 & 0.9072 & ±0.07 & 0.9013 & ±0.06 & 0.9020 & ±0.06 & 0.8456 & ±0.13 & 0.8471 & ±0.13 \\ \hline
 &  &  &  &  & CNN & 0.9786 & ±0.01 & 0.9790 & ±0.01 & 0.9584 & ±0.03 & 0.9594 & ±0.03 & 0.9443 & ±0.03 & 0.9451 & ±0.03 \\
 &  &  &  &  & RESNET18 & 0.9681 & ±0.02 & 0.9686 & ±0.02 & 0.9479 & ±0.03 & 0.9492 & ±0.02 & 0.9340 & ±0.03 & 0.9344 & ±0.03 \\
\multirow{-3}{*}{w/o mmd loss} & \multirow{-3}{*}{} & \multirow{-3}{*}{\checkmark} & \multirow{-3}{*}{\checkmark} & \multirow{-3}{*}{\checkmark} & TCN & 0.9496 & ±0.03 & 0.9506 & ±0.03 & 0.9093 & ±0.06 & 0.9095 & ±0.06 & 0.8650 & ±0.07 & 0.8618 & ±0.07 \\ \hline
 &  &  &  &  & CNN & 0.2012 & ±0.01 & 0.0558 & ±0.00 & 0.2028 & ±0.01 & 0.0562 & ±0.00 & 0.2039 & ±0.01 & 0.0564 & ±0.00 \\
 &  &  &  &  & RESNET18 & 0.2012 & ±0.01 & 0.0558 & ±0.00 & 0.2039 & ±0.01 & 0.0564 & ±0.00 & 0.2018 & ±0.01 & 0.0559 & ±0.00 \\
\multirow{-3}{*}{w/o ctr loss} & \multirow{-3}{*}{\checkmark} & \multirow{-3}{*}{} & \multirow{-3}{*}{} & \multirow{-3}{*}{\checkmark} & TCN & 0.2012 & ±0.01 & 0.0558 & ±0.00 & 0.2039 & ±0.01 & 0.0564 & ±0.00 & 0.2023 & ±0.01 & 0.0561 & ±0.00 \\ \hline
 &  &  &  &  & CNN & 0.9691 & ±0.02 & 0.9694 & ±0.02 & 0.8885 & ±0.11 & 0.8819 & ±0.13 & 0.8597 & ±0.11 & 0.8538 & ±0.12 \\
 &  &  &  &  & RESNET18 & 0.9488 & ±0.05 & 0.9424 & ±0.07 & 0.9463 & ±0.03 & 0.9465 & ±0.03 & 0.8875 & ±0.10 & 0.8813 & ±0.10 \\
\multirow{-3}{*}{w/o phase1 mix} & \multirow{-3}{*}{\checkmark} & \multirow{-3}{*}{\checkmark} & \multirow{-3}{*}{} & \multirow{-3}{*}{\checkmark} & TCN & 0.9162 & ±0.07 & 0.9176 & ±0.07 & 0.8524 & ±0.10 & 0.8535 & ±0.10 & 0.8031 & ±0.14 & 0.7980 & ±0.15 \\ \hline
 &  &  &  &  & CNN & 0.7973 & ±0.05 & 0.7942 & ±0.06 & 0.5713 & ±0.12 & 0.5429 & ±0.14 & 0.4805 & ±0.11 & 0.4479 & ±0.12 \\
 &  &  &  &  & RESNET18 & 0.5608 & ±0.06 & 0.5494 & ±0.06 & 0.4371 & ±0.08 & 0.4212 & ±0.09 & 0.3642 & ±0.06 & 0.3412 & ±0.06 \\
\multirow{-3}{*}{w/o 2-stage learning} & \multirow{-3}{*}{\checkmark} & \multirow{-3}{*}{\checkmark} & \multirow{-3}{*}{\checkmark} & \multirow{-3}{*}{} & TCN & 0.6048 & ±0.11 & 0.6008 & ±0.11 & 0.4055 & ±0.10 & 0.3950 & ±0.10 & 0.3421 & ±0.10 & 0.3266 & ±0.10 \\ \hline
\end{tabular}%
}
\caption{Ablation study on effectiveness of proposed methods, evaluated on HHAR dataset across domain pairs.}
\label{ablationHHAR_full}
\end{table*}

\begin{table*}[]
\centering
\resizebox{\textwidth}{!}{%
\begin{tabular}{cccccc|cccccccccccc}
\hline \hline
\multicolumn{6}{c}{} & \multicolumn{4}{c}{\rule{0pt}{13pt}\textbf{unlab.\_ratio   = 0.7}} & \multicolumn{4}{c}{\textbf{unlab.\_ratio = 0.9}} & \multicolumn{4}{c}{\textbf{unlab.\_ratio = 0.95}} \\
\multicolumn{6}{c}{\multirow{-2}{*}{}} & \multicolumn{2}{c}{accuracy} & \multicolumn{2}{c}{f1-score} & \multicolumn{2}{c}{accuracy} & \multicolumn{2}{c}{f1-score} & \multicolumn{2}{c}{accuracy} & \multicolumn{2}{c}{f1-score} \\
 & mmd\_loss & ctr\_loss & phase1 mix & 2-step & backbone & Avg. & std. & Avg. & std. & Avg. & std. & Avg. & std. & Avg. & std. & Avg. & std. \\ \hline
\rowcolor[gray]{0.9} 
\cellcolor[gray]{0.9} & \cellcolor[gray]{0.9} & \cellcolor[gray]{0.9} & \cellcolor[gray]{0.9} & \cellcolor[gray]{0.9} & CNN & 0.7294 & ±0.03 & 0.5962 & ±0.08 & 0.7209 & ±0.03 & 0.5719 & ±0.07 & 0.7005 & ±0.03 & 0.5632 & ±0.08 \\
\rowcolor[gray]{0.9} 
\cellcolor[gray]{0.9} & \cellcolor[gray]{0.9} & \cellcolor[gray]{0.9} & \cellcolor[gray]{0.9} & \cellcolor[gray]{0.9} & RESNET18 & 0.7361 & ±0.04 & 0.6179 & ±0.07 & 0.7213 & ±0.02 & 0.5880 & ±0.04 & 0.7014 & ±0.01 & 0.5701 & ±0.05 \\
\rowcolor[gray]{0.9} 
\multirow{-3}{*}{\cellcolor[gray]{0.9}Proposed} & \multirow{-3}{*}{\cellcolor[gray]{0.9}\checkmark} & \multirow{-3}{*}{\cellcolor[gray]{0.9}\checkmark} & \multirow{-3}{*}{\cellcolor[gray]{0.9}\checkmark} & \multirow{-3}{*}{\cellcolor[gray]{0.9}\checkmark} & TCN & 0.4284 & ±0.03 & 0.2551 & ±0.04 & 0.4389 & ±0.03 & 0.2459 & ±0.03 & 0.4451 & ±0.04 & 0.2299 & ±0.03 \\ \hline
 &  &  &  &  & CNN & 0.7397 & ±0.03 & 0.6098 & ±0.07 & 0.7029 & ±0.01 & 0.5490 & ±0.06 & 0.7088 & ±0.03 & 0.5525 & ±0.07 \\
 &  &  &  &  & RESNET18 & 0.7338 & ±0.04 & 0.6071 & ±0.08 & 0.7184 & ±0.02 & 0.5825 & ±0.06 & 0.6976 & ±0.02 & 0.5645 & ±0.05 \\
\multirow{-3}{*}{w/o mmd loss} & \multirow{-3}{*}{} & \multirow{-3}{*}{\checkmark} & \multirow{-3}{*}{\checkmark} & \multirow{-3}{*}{\checkmark} & TCN & 0.4519 & ±0.03 & 0.2433 & ±0.04 & 0.4451 & ±0.03 & 0.2424 & ±0.03 & 0.4515 & ±0.04 & 0.2361 & ±0.01 \\ \hline
 &  &  &  &  & CNN & 0.5147 & ±0.09 & 0.1352 & ±0.02 & 0.5147 & ±0.09 & 0.1352 & ±0.02 & 0.5147 & ±0.09 & 0.1352 & ±0.02 \\
 &  &  &  &  & RESNET18 & 0.5147 & ±0.09 & 0.1352 & ±0.02 & 0.5147 & ±0.09 & 0.1352 & ±0.02 & 0.5147 & ±0.09 & 0.1352 & ±0.02 \\
\multirow{-3}{*}{w/o ctr loss} & \multirow{-3}{*}{\checkmark} & \multirow{-3}{*}{} & \multirow{-3}{*}{} & \multirow{-3}{*}{\checkmark} & TCN & 0.5147 & ±0.09 & 0.1352 & ±0.02 & 0.5147 & ±0.09 & 0.1352 & ±0.02 & 0.5147 & ±0.09 & 0.1352 & ±0.02 \\ \hline
 &  &  &  &  & CNN & 0.7213 & ±0.04 & 0.5790 & ±0.08 & 0.7070 & ±0.05 & 0.5584 & ±0.07 & 0.6921 & ±0.05 & 0.5475 & ±0.08 \\
 &  &  &  &  & RESNET18 & 0.7188 & ±0.03 & 0.5755 & ±0.08 & 0.7032 & ±0.02 & 0.5585 & ±0.07 & 0.6828 & ±0.02 & 0.5273 & ±0.06 \\
\multirow{-3}{*}{w/o phase1 mix} & \multirow{-3}{*}{\checkmark} & \multirow{-3}{*}{\checkmark} & \multirow{-3}{*}{} & \multirow{-3}{*}{\checkmark} & TCN & 0.4521 & ±0.03 & 0.2570 & ±0.04 & 0.4507 & ±0.04 & 0.2431 & ±0.02 & 0.4456 & ±0.04 & 0.2360 & ±0.02 \\ \hline
 &  &  &  &  & CNN & 0.5651 & ±0.03 & 0.4239 & ±0.05 & 0.4802 & ±0.13 & 0.3354 & ±0.09 & 0.5341 & ±0.09 & 0.3788 & ±0.07 \\
 &  &  &  &  & RESNET18 & 0.6457 & ±0.04 & 0.5041 & ±0.05 & 0.5988 & ±0.04 & 0.4351 & ±0.03 & 0.5876 & ±0.05 & 0.4195 & ±0.04 \\
\multirow{-3}{*}{w/o 2-stage learning} & \multirow{-3}{*}{\checkmark} & \multirow{-3}{*}{\checkmark} & \multirow{-3}{*}{\checkmark} & \multirow{-3}{*}{} & TCN & 0.4158 & ±0.02 & 0.2469 & ±0.02 & 0.3914 & ±0.03 & 0.2286 & ±0.01 & 0.3986 & ±0.05 & 0.2246 & ±0.02 \\ \hline \hline
\end{tabular}%
}
\caption{Ablation study on effectiveness of proposed methods, evaluated on PTBXL dataset across domain pairs.}
\label{ablationPTBXL_full}
\end{table*}

\begin{table*}[]
\centering
\resizebox{\textwidth}{!}{%
\begin{tabular}{cccccc|cccccccccccc}
\hline
\multicolumn{6}{c|}{} & \multicolumn{4}{c}{\textbf{unlab.\_ratio   = 0.7}} & \multicolumn{4}{c}{\textbf{unlab.\_ratio = 0.9}} & \multicolumn{4}{c}{\textbf{unlab.\_ratio = 0.95}} \\
\multicolumn{6}{c|}{\multirow{-2}{*}{\textbf{}}} & \multicolumn{2}{c}{\textbf{accuracy}} & \multicolumn{2}{c}{\textbf{f1-score}} & \multicolumn{2}{c}{\textbf{accuracy}} & \multicolumn{2}{c}{\textbf{f1-score}} & \multicolumn{2}{c}{\textbf{accuracy}} & \multicolumn{2}{c}{\textbf{f1-score}} \\ \cline{7-18} 
 & mmd\_loss & ctr\_loss & phase1 mix & 2-step & backbone & Avg. & std. & Avg. & std. & Avg. & std. & Avg. & std. & Avg. & std. & Avg. & std. \\ \hline
\rowcolor[gray]{0.9} 
\cellcolor[gray]{0.9} & \cellcolor[gray]{0.9} & \cellcolor[gray]{0.9} & \cellcolor[gray]{0.9} & \cellcolor[gray]{0.9} & CNN & 0.8369 & ±0.02 & 0.7555 & ±0.04 & 0.8057 & ±0.04 & 0.6991 & ±0.05 & 0.7813 & ±0.05 & 0.5632 & ±0.08 \\
\rowcolor[gray]{0.9} 
\cellcolor[gray]{0.9} & \cellcolor[gray]{0.9} & \cellcolor[gray]{0.9} & \cellcolor[gray]{0.9} & \cellcolor[gray]{0.9} & RESNET18 & 0.7910 & ±0.05 & 0.6862 & ±0.05 & 0.7553 & ±0.07 & 0.6244 & ±0.06 & 0.7328 & ±0.07 & 0.5701 & ±0.05 \\
\rowcolor[gray]{0.9} 
\multirow{-3}{*}{\cellcolor[gray]{0.9}Proposed} & \multirow{-3}{*}{\cellcolor[gray]{0.9}\checkmark} & \multirow{-3}{*}{\cellcolor[gray]{0.9}\checkmark} & \multirow{-3}{*}{\cellcolor[gray]{0.9}\checkmark} & \multirow{-3}{*}{\cellcolor[gray]{0.9}\checkmark} & TCN & 0.4863 & ±0.04 & 0.3739 & ±0.05 & 0.4803 & ±0.06 & 0.3597 & ±0.05 & 0.4695 & ±0.05 & 0.2299 & ±0.03 \\ \hline
 &  &  &  &  & CNN & 0.8137 & ±0.05 & 0.7302 & ±0.06 & 0.7820 & ±0.06 & 0.6921 & ±0.07 & 0.7721 & ±0.07 & 0.5525 & ±0.07 \\
 &  &  &  &  & RESNET18 & 0.7917 & ±0.06 & 0.6886 & ±0.07 & 0.7734 & ±0.07 & 0.6439 & ±0.07 & 0.7437 & ±0.07 & 0.5645 & ±0.05 \\
\multirow{-3}{*}{w/o mmd loss} & \multirow{-3}{*}{} & \multirow{-3}{*}{\checkmark} & \multirow{-3}{*}{\checkmark} & \multirow{-3}{*}{\checkmark} & TCN & 0.5110 & ±0.07 & 0.3938 & ±0.05 & 0.5113 & ±0.05 & 0.3694 & ±0.06 & 0.5043 & ±0.05 & 0.2361 & ±0.01 \\ \hline
 &  &  &  &  & CNN & 0.4160 & ±0.06 & 0.1171 & ±0.01 & 0.4160 & ±0.06 & 0.1171 & ±0.01 & 0.4160 & ±0.06 & 0.1352 & ±0.02 \\
 &  &  &  &  & RESNET18 & 0.4160 & ±0.06 & 0.1171 & ±0.01 & 0.4160 & ±0.06 & 0.1171 & ±0.01 & 0.4160 & ±0.06 & 0.1352 & ±0.02 \\
\multirow{-3}{*}{w/o ctr loss} & \multirow{-3}{*}{\checkmark} & \multirow{-3}{*}{} & \multirow{-3}{*}{} & \multirow{-3}{*}{\checkmark} & TCN & 0.4160 & ±0.06 & 0.1171 & ±0.01 & 0.4160 & ±0.06 & 0.1171 & ±0.01 & 0.4160 & ±0.06 & 0.1352 & ±0.02 \\ \hline
 &  &  &  &  & CNN & 0.8245 & ±0.03 & 0.7421 & ±0.03 & 0.7679 & ±0.04 & 0.6472 & ±0.04 & 0.7392 & ±0.03 & 0.5475 & ±0.08 \\
 &  &  &  &  & RESNET18 & 0.7808 & ±0.05 & 0.6633 & ±0.04 & 0.7398 & ±0.07 & 0.5970 & ±0.08 & 0.7123 & ±0.09 & 0.5273 & ±0.06 \\
\multirow{-3}{*}{w/o phase1 mix} & \multirow{-3}{*}{\checkmark} & \multirow{-3}{*}{\checkmark} & \multirow{-3}{*}{} & \multirow{-3}{*}{\checkmark} & TCN & 0.4825 & ±0.03 & 0.3609 & ±0.05 & 0.4560 & ±0.05 & 0.3332 & ±0.05 & 0.4495 & ±0.05 & 0.2360 & ±0.02 \\ \hline
 &  &  &  &  & CNN & 0.6464 & ±0.10 & 0.5570 & ±0.09 & 0.6232 & ±0.08 & 0.4596 & ±0.10 & 0.5557 & ±0.11 & 0.3788 & ±0.07 \\
 &  &  &  &  & RESNET18 & 0.6704 & ±0.04 & 0.5638 & ±0.04 & 0.5851 & ±0.06 & 0.4510 & ±0.07 & 0.5538 & ±0.07 & 0.4195 & ±0.04 \\
\multirow{-3}{*}{w/o 2-stage learning} & \multirow{-3}{*}{\checkmark} & \multirow{-3}{*}{\checkmark} & \multirow{-3}{*}{\checkmark} & \multirow{-3}{*}{} & TCN & 0.4252 & ±0.05 & 0.3363 & ±0.05 & 0.3807 & ±0.05 & 0.3058 & ±0.05 & 0.3992 & ±0.05 & 0.2246 & ±0.02 \\ \hline
\end{tabular}%
}
\caption{Ablation study on effectiveness of proposed methods, evaluated on EEG dataset across domain pairs.}
\label{ablationEEG_full}
\end{table*}

\begin{table*}[]
\centering
\resizebox{\textwidth}{!}{%
\begin{tabular}{cccccc|cccccccccccc}
\hline
\multicolumn{6}{c|}{} & \multicolumn{4}{c}{\textbf{unlab.\_ratio   = 0.7}} & \multicolumn{4}{c}{\textbf{unlab.\_ratio = 0.9}} & \multicolumn{4}{c}{\textbf{unlab.\_ratio = 0.95}} \\
\multicolumn{6}{c|}{\multirow{-2}{*}{\textbf{}}} & \multicolumn{2}{c}{\textbf{accuracy}} & \multicolumn{2}{c}{\textbf{f1-score}} & \multicolumn{2}{c}{\textbf{accuracy}} & \multicolumn{2}{c}{\textbf{f1-score}} & \multicolumn{2}{c}{\textbf{accuracy}} & \multicolumn{2}{c}{\textbf{f1-score}} \\ \cline{7-18} 
 & mmd\_loss & ctr\_loss & phase1 mix & 2-step & backbone & Avg. & std. & Avg. & std. & Avg. & std. & Avg. & std. & Avg. & std. & Avg. & std. \\ \hline
\rowcolor[gray]{0.9} 
\cellcolor[gray]{0.9} & \cellcolor[gray]{0.9} & \cellcolor[gray]{0.9} & \cellcolor[gray]{0.9} & \cellcolor[gray]{0.9} & CNN & 0.9832 & ±0.03 & 0.9793 & ±0.03 & 0.9777 & ±0.05 & 0.9759 & ±0.05 & 0.9798 & ±0.03 & 0.9736 & ±0.03 \\
\rowcolor[gray]{0.9} 
\cellcolor[gray]{0.9} & \cellcolor[gray]{0.9} & \cellcolor[gray]{0.9} & \cellcolor[gray]{0.9} & \cellcolor[gray]{0.9} & RESNET18 & 0.9726 & ±0.04 & 0.9571 & ±0.07 & 0.9339 & ±0.11 & 0.9065 & ±0.15 & 0.9519 & ±0.06 & 0.9096 & ±0.14 \\
\rowcolor[gray]{0.9} 
\multirow{-3}{*}{\cellcolor[gray]{0.9}Proposed} & \multirow{-3}{*}{\cellcolor[gray]{0.9}\checkmark} & \multirow{-3}{*}{\cellcolor[gray]{0.9}\checkmark} & \multirow{-3}{*}{\cellcolor[gray]{0.9}\checkmark} & \multirow{-3}{*}{\cellcolor[gray]{0.9}\checkmark} & TCN & 0.6062 & ±0.03 & 0.6225 & ±0.04 & 0.5923 & ±0.03 & 0.6062 & ±0.04 & 0.5828 & ±0.03 & 0.6052 & ±0.04 \\ \hline
 &  &  &  &  & CNN & 0.9779 & ±0.05 & 0.9668 & ±0.07 & 0.9651 & ±0.05 & 0.9246 & ±0.13 & 0.9710 & ±0.04 & 0.9172 & ±0.14 \\
 &  &  &  &  & RESNET18 & 0.9687 & ±0.05 & 0.9363 & ±0.11 & 0.9581 & ±0.06 & 0.9306 & ±0.10 & 0.9448 & ±0.06 & 0.8661 & ±0.16 \\
\multirow{-3}{*}{w/o mmd loss} & \multirow{-3}{*}{} & \multirow{-3}{*}{\checkmark} & \multirow{-3}{*}{\checkmark} & \multirow{-3}{*}{\checkmark} & TCN & 0.6138 & ±0.04 & 0.6265 & ±0.06 & 0.6060 & ±0.03 & 0.6211 & ±0.05 & 0.5929 & ±0.03 & 0.6127 & ±0.05 \\ \hline
 &  &  &  &  & CNN & 0.4550 & ±0.00 & 0.2085 & ±0.00 & 0.4539 & ±0.00 & 0.2081 & ±0.00 & 0.4550 & ±0.00 & 0.2085 & ±0.00 \\
 &  &  &  &  & RESNET18 & 0.4539 & ±0.00 & 0.2081 & ±0.00 & 0.4539 & ±0.00 & 0.2081 & ±0.00 & 0.4539 & ±0.00 & 0.2081 & ±0.00 \\
\multirow{-3}{*}{w/o ctr loss} & \multirow{-3}{*}{\checkmark} & \multirow{-3}{*}{} & \multirow{-3}{*}{} & \multirow{-3}{*}{\checkmark} & TCN & 0.4539 & ±0.00 & 0.2081 & ±0.00 & 0.4539 & ±0.00 & 0.2081 & ±0.00 & 0.4539 & ±0.00 & 0.2081 & ±0.00 \\ \hline
 &  &  &  &  & CNN & 0.9661 & ±0.03 & 0.9565 & ±0.04 & 0.8842 & ±0.08 & 0.8449 & ±0.15 & 0.8635 & ±0.11 & 0.8472 & ±0.15 \\
 &  &  &  &  & RESNET18 & 0.9797 & ±0.03 & 0.9792 & ±0.03 & 0.9171 & ±0.08 & 0.8998 & ±0.12 & 0.8912 & ±0.09 & 0.8992 & ±0.08 \\
\multirow{-3}{*}{w/o phase1 mix} & \multirow{-3}{*}{\checkmark} & \multirow{-3}{*}{\checkmark} & \multirow{-3}{*}{} & \multirow{-3}{*}{\checkmark} & TCN & 0.5853 & ±0.03 & 0.5958 & ±0.05 & 0.5832 & ±0.02 & 0.5875 & ±0.05 & 0.5713 & ±0.03 & 0.5741 & ±0.05 \\ \hline
 &  &  &  &  & CNN & 0.8632 & ±0.16 & 0.8098 & ±0.23 & 0.7862 & ±0.17 & 0.7604 & ±0.21 & 0.8063 & ±0.13 & 0.7805 & ±0.16 \\
 &  &  &  &  & RESNET18 & 0.7958 & ±0.15 & 0.7753 & ±0.16 & 0.7483 & ±0.15 & 0.6853 & ±0.18 & 0.7219 & ±0.19 & 0.6542 & ±0.20 \\
\multirow{-3}{*}{w/o 2-stage learning} & \multirow{-3}{*}{\checkmark} & \multirow{-3}{*}{\checkmark} & \multirow{-3}{*}{\checkmark} & \multirow{-3}{*}{} & TCN & 0.5408 & ±0.02 & 0.5220 & ±0.04 & 0.5266 & ±0.02 & 0.4956 & ±0.04 & 0.5045 & ±0.03 & 0.4541 & ±0.05 \\ \hline
\end{tabular}%
}
\caption{Ablation study on effectiveness of proposed methods, evaluated on MFD dataset across domain pairs.}
\label{ablationMFD_full}
\end{table*}

\suppsubsection{Performance Details on Full Scenarios}
Table \ref{detailPTBXL} summarizes the complete cross-domain performance of MoSSDA on the PTBXL dataset. Since PTBXL defines three domains, all six possible source-target pairings were evaluated, allowing for a comprehensive view of generalizability across domain shifts.
Similarly, Table \ref{detailWISDM} reports MoSSDA’s performance on 10 selected source-target domain pairs out of the 36 defined combinations in the WISDM dataset. These results collectively assess the scalability and domain transferability of our proposed method under realistic and diverse deployment conditions.
\begin{table*}[]
\centering
\resizebox{0.7\textwidth}{!}{%
\begin{tabular}{cccccccc}
\hline\hline
 &  & \multicolumn{6}{c}{\textbf{\rule{0pt}{13pt}Scenario ( T to S )}} \\
unlab.\_ratio & \multicolumn{1}{c|}{} & 1\_to\_2 & 1\_to\_3 & 2\_to\_1 & 2\_to\_3 & 3\_to\_1 & 3\_to\_2 \\ \hline
\multirow{3}{*}{0.7}  & \multicolumn{1}{c|}{accuracy}  & 0.7263   & 0.7222                       & 0.7727                       & 0.7440                       & 0.7683                       & 0.6733                       \\
                      & \multicolumn{1}{c|}{f1\_score} & 0.5883   & 0.5517                       & 0.6963                       & 0.6038                       & 0.6796                       & 0.5582                       \\
                      & \multicolumn{1}{c|}{auorc}     & 0.8194   & 0.8565                       & 0.8991                       & 0.8384                       & 0.8956                       & 0.8232                       \\ \hline
\multirow{3}{*}{0.9}  & \multicolumn{1}{c|}{accuracy}  & 0.6813   & 0.7076                       & 0.7098                       & 0.7409                       & 0.7507                       & 0.6973                       \\
                      & \multicolumn{1}{c|}{f1\_score} & 0.5503   & 0.5307                       & 0.6000                       & 0.5840                       & 0.6460                       & 0.5791                       \\
                      & \multicolumn{1}{c|}{auorc}     & 0.8304   & 0.8335                       & 0.8546                       & 0.8304                       & 0.8744                       & 0.8335                       \\ \hline
\multirow{3}{*}{0.95} & \multicolumn{1}{c|}{accuracy}  & 0.6953   & 0.6837                       & 0.7032                       & 0.7055                       & 0.7135                       & 0.7013                       \\
                      & \multicolumn{1}{c|}{f1\_score} & 0.5600   & 0.5032                       & 0.6052                       & 0.5337                       & 0.6220                       & 0.5865                       \\
                      & \multicolumn{1}{c|}{auorc}     & 0.8170   & 0.8143                       & 0.8673                       & 0.7745                       & 0.8552                       & 0.8160                       \\ \hline\hline
\end{tabular}%
}
\caption{MoSSDA performances are evaluated on PTBXL dataset in all possible domain scenario and 3 fixed unlabeled ratio using RESNET18 as backbone.  }
\label{detailPTBXL}
\end{table*}

\begin{table*}[]
\centering
\resizebox{\textwidth}{!}{%
\begin{tabular}{ccccccccccc}
\hline\hline
 &  & \multicolumn{9}{c}{\textbf{\rule{0pt}{13pt}Scenario ( T to S )}} \\
unlab. ratio & \multicolumn{1}{c|}{} & 20\_to\_30 & 23\_to\_32 & 28\_to\_4 & 2\_to\_11 & 33\_to\_12 & 35\_to\_31 & 5\_to\_26 & 6\_to\_19 & 7\_to\_18 \\ \hline
\multirow{3}{*}{0.7}  & \multicolumn{1}{c|}{accuracy}  & 0.8350     & 0.7826     & 0.8789    & 0.7368    & 0.7931     & 0.8193     & 0.8659    & 0.8788    & 0.7925    \\
                      & \multicolumn{1}{c|}{f1\_score} & 0.7093     & 0.7201     & 0.8268    & 0.5294    & 0.5047     & 0.7461     & 0.8165    & 0.8087    & 0.6563    \\
                      & \multicolumn{1}{c|}{auorc}     & 0.9405     & 0.9171     & 0.9570    & 0.9080    & 0.9730     & 0.9438     & 0.9301    & 0.956     & 0.8204    \\ \hline
\multirow{3}{*}{0.9}  & \multicolumn{1}{c|}{accuracy}  & 0.7670     & 0.8116     & 0.8030    & 0.7763    & 0.6437     & 0.8675     & 0.8049    & 0.8106    & 0.7453    \\
                      & \multicolumn{1}{c|}{f1\_score} & 0.6456     & 0.7369     & 0.7648    & 0.6826    & 0.4112     & 0.7335     & 0.6175    & 0.6618    & 0.6305    \\
                      & \multicolumn{1}{c|}{auorc}     & 0.9076     & 0.8767     & 0.9221    & 0.8715    & 0.7011     & 0.9603     & 0.8879    & 0.9437    & 0.7577    \\ \hline
\multirow{3}{*}{0.95} & \multicolumn{1}{c|}{accuracy}  & 0.6893     & 0.6667     & 0.7879    & 0.6447    & 0.6437     & 0.6506     & 0.7683    & 0.7879    & 0.6887    \\
                      & \multicolumn{1}{c|}{f1\_score} & 0.6076     & 0.6461     & 0.7245    & 0.6468    & 0.4841     & 0.3902     & 0.5496    & 0.7253    & 0.5746    \\
                      & \multicolumn{1}{c|}{auorc}     & 0.8940     & 0.8360     & 0.9448    & 0.8077    & 0.8185     & 0.8927     & 0.8738    & 0.9173    & 0.7509    \\ \hline\hline
\end{tabular}%
}
\caption{MoSSDA performances are evaluated on WISDM dataset in randomly fixed 10 domain scenario and 3 fixed unlabeled ratio using TCN as backbone.  }
\label{detailWISDM}
\end{table*}

\end{document}